\documentclass[runningheads]{llncs}

\usepackage{eccv}

\usepackage{eccvabbrv}

\usepackage{graphicx}
\usepackage{booktabs}

\usepackage{xcolor,pifont}
\usepackage{amsmath}
\usepackage{bbm}
\usepackage{multirow}
\usepackage{makecell}
\usepackage{adjustbox}

\definecolor{coolblack}{rgb}{0.184, 0.33, 0.59}
\definecolor{newpink}{rgb}{0.984, 0.616, 0.741}
\definecolor{foregreen}{rgb}{0.133,0.545,0,133}
\definecolor{babyblue}{rgb}{0.54, 0.81, 0.94}
\definecolor{deeppeach}{rgb}{1.0, 0.8, 0.64}

\usepackage[accsupp]{axessibility}  

\usepackage{hyperref}

\usepackage{orcidlink}

\begin{document}

\title{SelfSwapper: Self-Supervised Face Swapping via Shape Agnostic Masked AutoEncoder} 

\titlerunning{SelfSwapper}

\author{Jaeseong Lee\inst{*} \and
Junha Hyung\inst{*} \and
Sohyun Jung \and
Jaegul Choo 
}
\authorrunning{J. Lee et al.}

\institute{KAIST AI
\\
\email{\{wintermad1245, sharpeeee, jsh0212, jchoo\}@kaist.ac.kr 
} \\
* indicates equal contributions.}
\maketitle
\begin{center}
    \href{https://summertight.github.io/selfswapper/}{\textcolor{coolblack}{summertight.github.io/selfswapper}}
\end{center}

\begin{figure}
    \centering 
    
    \includegraphics[width=1.\linewidth]{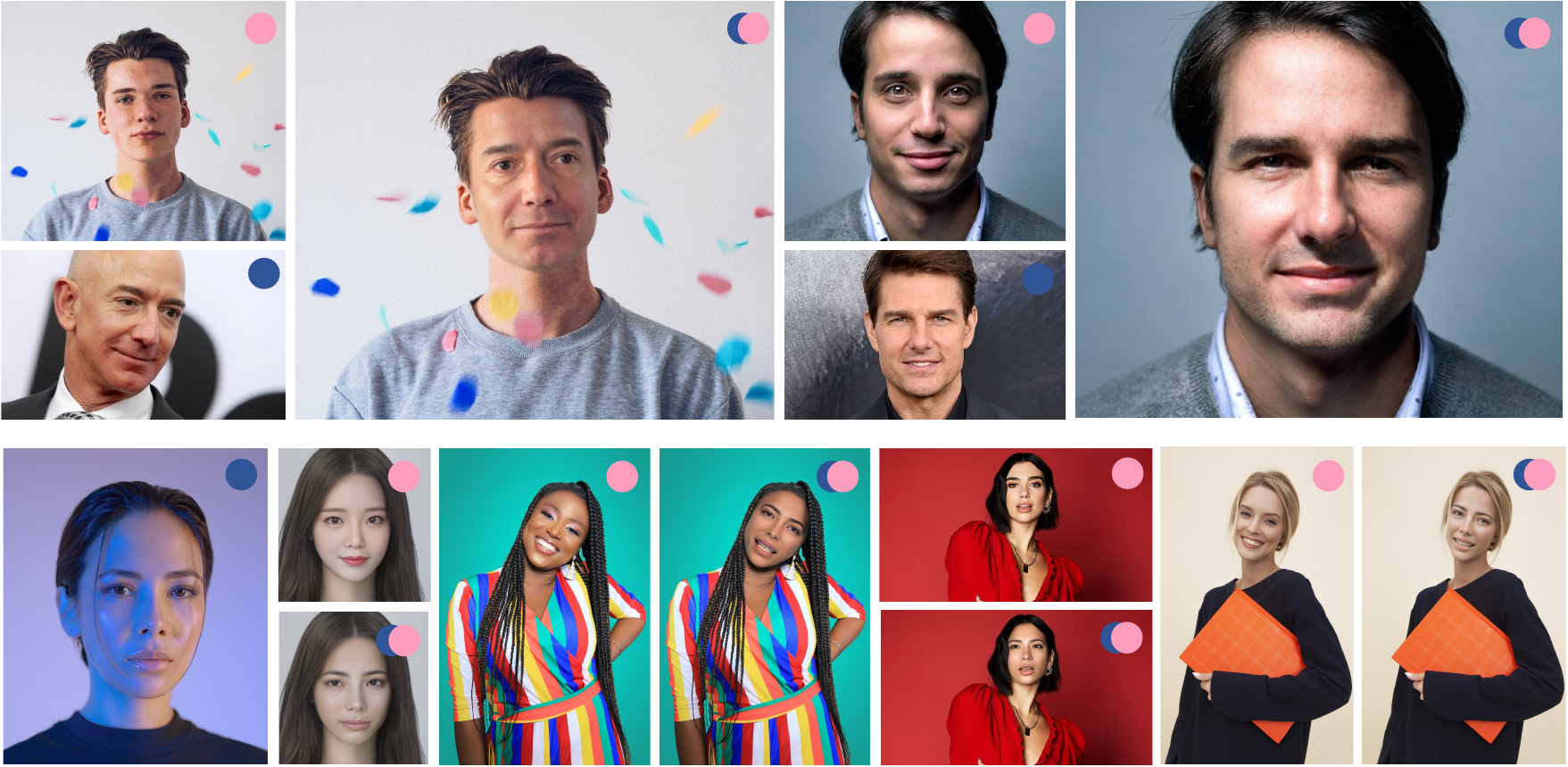}
    \caption{\textbf{Real-world application of our model.} This figure showcases results with the in-the-wild samples, \textcolor{coolblack}{navy} circles for the source, \textcolor{newpink}{pink} circles for the target, and overlaps showing the generated outputs. The second row displays one-source multi-target results. Our model accurately transforms the target face to match the source while faithfully preserving the target attributes such as the skin color, pose, expression, hair, background, and gaze. This showcases our model's robustness on in-the-wild samples and real-world applicability for diverse facial images. For resolutions beyond $256\times256$, an off-the-shelf super-resolution model~\cite{restoreformer} is used.}
\label{fig:teaser}
\vspace{-1.1cm}
\end{figure}


\begin{abstract}
Face swapping has gained significant attention for its varied applications.
Most previous face swapping approaches have relied on the seesaw game training scheme, also known as the target-oriented approach. 
However, this often leads to instability in model training and results in undesired samples with blended identities due to the target identity leakage problem.
Source-oriented methods achieve more stable training with self-reconstruction objective but often fail to accurately reflect target image's skin color and illumination.
This paper introduces the Shape Agnostic Masked AutoEncoder (SAMAE) training scheme, a novel self-supervised approach that combines the strengths of both target-oriented and source-oriented approaches.
Our training scheme addresses the limitations of traditional training methods by circumventing the conventional seesaw game and introducing clear ground truth through its self-reconstruction training regime.
Our model effectively mitigates identity leakage and reflects target albedo and illumination through learned disentangled identity and non-identity features. 
Additionally, we closely tackle the shape misalignment and volume discrepancy problems with new techniques, including perforation confusion and random mesh scaling.
SAMAE establishes a new state-of-the-art, surpassing other baseline methods, preserving both identity and non-identity attributes without sacrificing on either aspect. 

\end{abstract}


\section{Introduction}
\label{sec:intro}

Face swapping has gained significant attention for its ability to create virtual human avatars, digitally resurrect individuals, and develop virtual models. 
This growing interest is matched by extensive research in academia, driven by advancements in deep generative models~\cite{stylegan2,gan,vae,ddpm}. 
Face swapping aims to seamlessly merge the identity characteristics of a source face with the non-identity features (\eg, skin tone, pose, and lighting) of a target face, thereby crafting a cohesive and realistic facial image.

The majority of previous approaches~\cite{simswap,blendface,smoothswap,robustswap, infoswap,hififace,faceshifter,reinforceswap,diffswap} often employ the \textit{seesaw game} training scheme, a multi-task learning strategy that leverages both reconstruction loss and identity loss~\cite{arcface} to achieve dual objectives: maintaining non-identity attributes of the target image and capturing identity of the source image.
Known as the \textbf{target-oriented} approach, this scheme is particularly effective in face swapping, where ground-truth images are unavailable.
This approach efficiently incorporates knowledge from pre-trained face recognition models~\cite{arcface} for reflecting identity information. 
Additionally, reconstruction losses such as L1 and Perceptual~\cite{vgg} ensure comprehensive preservation of the target image's attributes, such as skin color and illumination. 

However, training methods that depend on two loss kinds with distinct goals without clear ground truth often lead to instability and unreliability in the model.
The conflicting nature of these losses can lead to unstable gradient flow~\cite{smoothswap}, misleading the models to reflect imprecise identity information and reducing model consistency across different samples.
Simple reconstruction losses may cause unwanted mixing of source and target identities, resulting in the identity leakage from the target image.
Carefully tuned hyperparameters are needed to balance these competing objectives, which complicates the training process. 


On the contrary, the recent \textbf{source-oriented} approaches~\cite{e4s,fsgan} adopt a \sloppy{self-supervised framework}, incorporating a straightforward self-reconstruction task along with established head reenactment models~\cite{fomm,lia,osfv} to morph the source image into the target pose. It then seamlessly integrates the reenacted source image with the target background at the pixel level.
However, its self-supervised training strategy overlooks the cross-identity inference phase, leading to difficulties with generalization.
For instance, the model often transfers the color attributes of the source to the generated output, which, along with target body's heterogeneous skin tone and illumination, leads to unrealistic results.

Moreover, addressing shape and volume misalignment between the source and target face is crucial. The facial shape significantly influences facial identity, and precise alignment of facial volume is essential for achieving realistic results. Many previous methods overlook this aspect, and even those that consider it, such as DiffSwap~\cite{diffswap}, yield unsatisfactory outcomes due to vague guidance provided during contour generation in the seesaw game training scheme.

In this paper, we introduce the Shape Agnostic Masked AutoEncoder (SAMAE), a framework that combines the strengths of both target-oriented and source-oriented approaches.
The SAMAE employs self-supervised training, which introduces clear ground truth through its self-reconstruction training regime, effectively addressing the conventional seesaw game issue.
Additionally, it achieves high generalization performance including accurate incorporation of target skin color and illumination in the cross-view inference phase, where previous target-oriented methods have faltered.
SAMAE also adeptly handles shape and volume misalignment problems.

This generalizable self-supervised training is achieved through 1) the model design and 2) novel techniques including \textit{perforation confusion} and \textit{random mesh scaling}.
Our model features learnable modules that disentangle identity and non-identity features, along with a combination of pretrained models like 3D Morphable Models (3DMMs)~\cite{bfm,deep3d} and face identity encoders~\cite{arcface}. 
Our design also allows for accurate skin color and illumination extraction from the target.


We introduce perforation confusion and random mesh scaling to enhance the cross-identity inference capability.
Perforation confusion creates shape-agnostic masks during training, addressing the shape misalignment problem.
Random mesh scaling is designed to mitigate the model's excessive dependence on pixel-aligned information from the conditioned mesh, enabling the model to better manage volume discrepancies between the source and target faces.

\begin{figure}[tb]
  \centering
    \includegraphics[width=\linewidth]{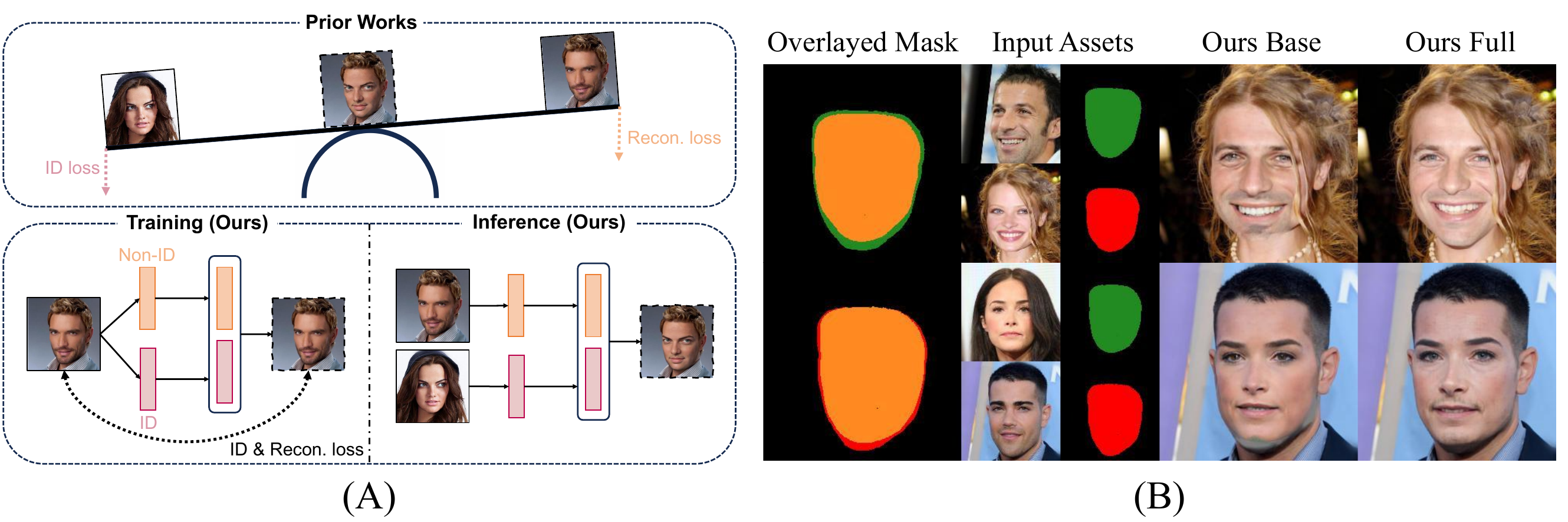}
    \caption{(A) \textbf{Conceptual comparison between prior works and our method.} Prior works rely on a seesaw game of two potentially conflicting losses: reconstruction loss and identity loss. On the other hand, our method leverages a self-supervised approach with a clear ground truth, which allows for more stable training. (B) \textbf{Comparing our base approach (Ours Base) with our enhanced method (Ours Full)}, which includes techniques like perforation confusion and random mesh scaling. \textcolor{foregreen}{Green masks} represent target-posed source 3DMM masks, \textcolor{red}{red masks} indicate target 3DMM masks, and \textcolor{orange}{orange masks} denote their intersection. The first row shows that when the source face is larger than the target's, the jaw is cut off. The second row shows the opposite case, where the base model fails to inpaint the remaining regions effectively, while Ours Full generates realistic face-swapped outputs.}

  \label{fig:concept}
\end{figure}

In summary, our contributions are:\\
\begin{itemize}
    \item We introduce a novel training regime of the Shape-Agnostic Masked AutoEncoder (SAMAE) in face swapping, which effectively eliminates the unstable \textit{seesaw game} and addresses the problem of target identity leakage through self-supervised learning, while disentangling target albedo and illumination.\\
    \item We address the shape-misalignment and facial volume discrepancy problem with two innovative yet straightforward techniques: \textit{perforation confusion} and \textit{random mesh scaling}. \\
    \item Our new training approach establishes a new state-of-the-art, surpassing other baseline methods in terms of both identity and non-identity attributes without sacrificing on either aspect.

\end{itemize}



\section{Related Work}
\noindent\textbf{Target-Oriented Face Swapping.}
The target-oriented face swapping refers to a group of face swapping methods that operate by manipulating the target spatial features and leveraging the off-the-shelf face recognition~\cite{arcface} models.
This approach inherently relies on seesaw game between reconstruction loss and identity loss.
DeepFakes~\cite{deepfakes} initially introduced an autoencoder-based algorithm capable of swapping faces between two specifically trained identities, but it lacked generalization capabilities.
Faceshifter~\cite{faceshifter} proposes a Adaptive Embedding Intergration Network (AEI-Net) for merging target non-identity attributes with source identity.
SimSwap~\cite{simswap} uses Weak Feature Matching (WFM) loss to maintain target attributes.
InfoSwap~\cite{infoswap} applies an information bottleneck principle to filter out target identity information. 
HifiFace~\cite{hififace} adopts a 3DMM parameter-based swapping method. 
SmoothSwap~\cite{smoothswap} focuses on identity space smoothness, improving source identity preservation at the cost of target attribute fidelity.
MegaFS~\cite{megafs} and FSLSD~\cite{fslsd} utilize StyleGAN for high-resolution outputs and structure disentanglement, respectively.
RobustSwap~\cite{robustswap} and BlendFace~\cite{blendface} address attribute leakage in face swapping.
DiffSwap~\cite{diffswap} tackles face swapping as an inpainting problem, leveraging the latent diffusion models~\cite{ldm}.
The work tackles the shape misalignment problem by using convex-hulls of keypoints from both source and target as perforation masks. However, it fails to explicitly disentangle facial contour information from the input conditions due to their seesaw training regime.
WSC-swap~\cite{reinforceswap} critiques the source-target disentanglement in existing methods. 
However, all these methods still engage in a seesaw game with the agenda of disentangling source-target information. In contrast, we propose a brand-new training paradigm for face swapping that is free from this seesaw game.
\\

\noindent\textbf{Source-Oriented Face Swapping.}
Face swapping methods with a source-oriented approach frequently integrate classical point mapping techniques~\cite{umeyama} or employ face reenactment models~\cite{osfv,fomm,lia} to warp the source image to match the pose of the target and blend the source face with target's background. 
DeepFaceLab~\cite{deepfacelab} belongs to the former camp, employing similarity transformation-based warping and pasting followed by sharpening and blending strategies in post-processing. These methods typically excludes the target's facial region during network training, free from target identity leakage. FSGAN~\cite{fsgan}, belonging to the latter category, integrates face swapping and reenactment in a two-stage process. This is followed by the use of a face inpainting network to blend the reenacted source with the target images.
The most recent work in the latter category, E4S~\cite{e4s}, introduced a regional GAN inversion approach along with face reenactment and mask editing.
However, these methods, dependent on reenactment models and point mapping methods, often struggle to capture the target's illumination and can result in identity shifts, particularly when there are large pose discrepancies between the source and the target. 
It arises from the fact that, despite their approaches originating from the self-supervised manner, their training schema is not enough to consider test phase scenarios.
Furthermore, they show inferior performance when the source image includes self-occlusions or accessories.
\\

\noindent\textbf{Face Disentanglement Learning.}
DiscoFaceGAN~\cite{discofacegan} disentangles latent space of StyleGAN~\cite{stylegan2} into 3DMM parameters for controlled generation. GIF~\cite{gif} combines 3DMM-rendered assets with StyleGAN's noise space to control expression and lighting. 
CONFIG~\cite{config} differentiates between real image latents and 3DMM parameters, while VariTex~\cite{varitex} uses variational space-based UV projection for manipulation. 
3DFM-GAN~\cite{3dfmgan} refines StyleGAN's latent space for controllable manipulation. 
However, these methods aggregate facial and non-facial elements into identity, complicating their use in face swapping. 




\begin{figure*}[t]
  \centering
   \includegraphics[width=\linewidth]{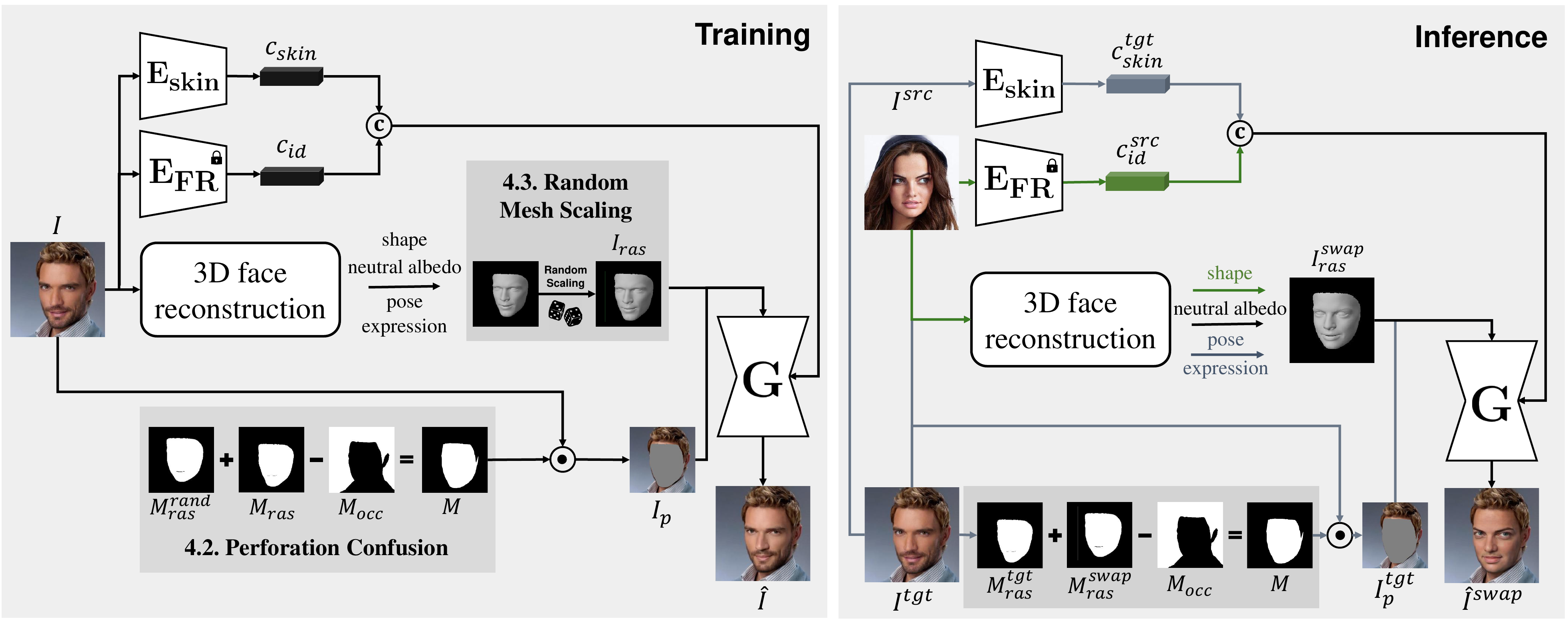}

   \caption{\textbf{Overall pipeline of our method.} In the training phase (left), we employ self-reconstruction scheme with perforation confusion and random mesh scaling, enhancing shape agnostic robustness for SAMAE training. During inference (right), this training enables the model to efficiently perform cross-identity face swapping by disentangling ID and non-ID attributes.}
   \label{fig:pipeline}
\end{figure*}

\section{Backgrounds}
\textbf{Essentials of Face Swapping.}
High-fidelity face-swapped results, $\hat{I}^\textit{swap}$, should adhere to the following conditions, inheriting identity from a source image $I^\textit{src}$ and non-identity attributes from a target image $I^\textit{tgt}$:
\begin{itemize}
    
    \item \textcolor{olive}{C1}. They should preserve the non-identity attributes of $I^\textit{tgt}$, which include non-facial attributes (\eg, background and hair), facial posture (\eg, expression and pose), and facial color (\eg, skin color and lighting). 
    \item \textcolor{olive}{C2}. They must maintain the identity features such as facial contour, inner facial traits (\eg, eyes, eyebrows, nose, lip, cheekbone, dimples, and their interrelations), and skin details (\eg, freckles, moles, and wrinkles) of the source image $I^\textit{src}$.
    \item \textcolor{olive}{C3}. The resulting images should be indistinguishable from real images.
\end{itemize}
\textbf{3D Morphable Model (3DMM).} We employ an off-the-shelf 3DMM~\cite{bfm} and 3DMM parameter estimator $E_\textit{3DMM}$~\cite{deep3d}, coupled for 3D face reconstruction. The $E_\textit{3DMM}$ estimates a tuple of the lighting and geometric parameters $v=(\alpha, \beta, \gamma, \delta, R, t_{xy}, s)$ of a facial image. The $v$ is composed of shape $\alpha$, expression $\beta$, albedo $\gamma$, lighting $\delta$, and camera parameters including rotation matrix $R$, xy-axial translation parameter $t_{xy}$ and object scale $s$.\footnote{In most 3DMMs, as they adopt an orthographic camera model, there is no z-axis translation $t_{z}$.} The renderer $Rd$ outputs the rasterized face mesh image $I_\textit{ras}=Rd(v)$. The foreground mesh mask $M_\textit{ras}$ is automatically derived from the $I_\textit{ras}$.

\section{Method}
\subsection{Shape Agnostic Masked AutoEncoder}  
The goal of the Shape Agnostic Masked AutoEncoder (SAMAE) training regime is to reconstruct the ground truth image $I$ given the corresponding decomposed components, which include identity (\textcolor{olive}{C2}) and non-identity attributes (\textcolor{olive}{C1}).
To extract the non-facial attributes from the image, we utilize a foreground mesh mask $M_\textit{ras}$ and an occlusion mask $M_\textit{occ}$, obtained from the-off-the-shelf face parser~\cite{bisenet}. This strategy is the most straightforward way to exclude the target facial identity information, aligning with the principles of source-oriented methods.
By removing the occluded region $M_\textit{occ}$ from $M_\textit{ras}$, we construct the final facial mask $M$ as $M = M_\textit{ras} - M_\textit{occ}$. 
The non-facial attribute image is then defined by $I_{p}  = (\mathbbm{1}-M)\odot I$, where $\odot$ represents the Hadamard product and $\mathbbm{1}$ is a 1-filled tensor matching the dimension of $M$.

The facial posture and facial color are represented by $(\beta, R)$ and $(c_\textit{skin}, \delta)$, respectively. $c_\textit{skin}$ is derived from the skin area of the image $I$ using the skin color encoder $E_\textit{skin}$ (Sec.~\ref{subsec:skin_color}).
For the identity, the facial contour can be expressed by the facial mesh outline of $I_\textit{ras}$, and the inner facial traits and the skin details are represented by the combination of the shape parameter $\alpha$ and identity embedding $c_\textit{id}$, where $c_\textit{id}$ is extracted from the pretrained face recognition model $E_\textit{FR}$. 

These features are conditioned to the U-Net based generator $G$~\cite{adm} which outputs the reconstructed image $\hat{I}$:
\begin{equation}
\hat{I} = G(I_\textit{ras},I_{p},c_\textit{id},c_\textit{skin}) = G(Rd(v), I_\textit{p},E_\textit{FR}(I),E_\textit{skin}(I)),
\end{equation}
where $v$ here is an estimated 3DMM parameter with albedo $\gamma$ replaced to a neutralized albedo $\gamma_\textit{neu}$, which will be discussed in Sec.~\ref{subsec:skin_color}.

Switching to the cross-identity inference (swap) regime, we condition the model with the identity-relevant information extracted from the source image $I^\textit{src}$ to generate the swapped result $\hat{I}^\textit{swap}$,
\begin{equation}
\begin{split}
\hat{I}^\textit{swap} & = G(I_\textit{ras}^\textit{swap}, I_\textit{p}^\textit{tgt}, c_\textit{id}^\textit{src}, c_\textit{skin}^\textit{tgt}) \\
              & = G(Rd(v^\textit{swap}), I_\textit{p}^\textit{tgt},E_\textit{FR}(I^\textit{src}),E_\textit{skin}(I^\textit{tgt})), 
\end{split}
\end{equation}
where $v^\textit{swap} = (\alpha^\textit{src}, \beta^\textit{tgt}, \gamma_\textit{neu}, \delta^\textit{tgt}, R^\textit{tgt}, t_\textit{xy}^\textit{tgt}, s^\textit{tgt})$ and superscripts \textit{src} and \textit{tgt} indicates whether the features are from $I^\textit{src}$ or $I^\textit{tgt}$.
Furthermore, since 3DMMs offer limited representation for eye-gazing, we leverage the positions of the target image's iris keypoints from~\cite{iris}.
Keypoints are represented as a stickmen-like rendering (as described in~\cite{fsth}), and we utilize it as a spatial input of the model, concatenating it with $I_\textit{ras}$ and $I_\textit{p}$. 
For brevity, we omit notions of this input in this manuscript.
For overall pipeline of training and inference, please refer to Fig.~\ref{fig:pipeline}. 

\begin{figure*}[t]
  \centering
   \includegraphics[width=\linewidth]{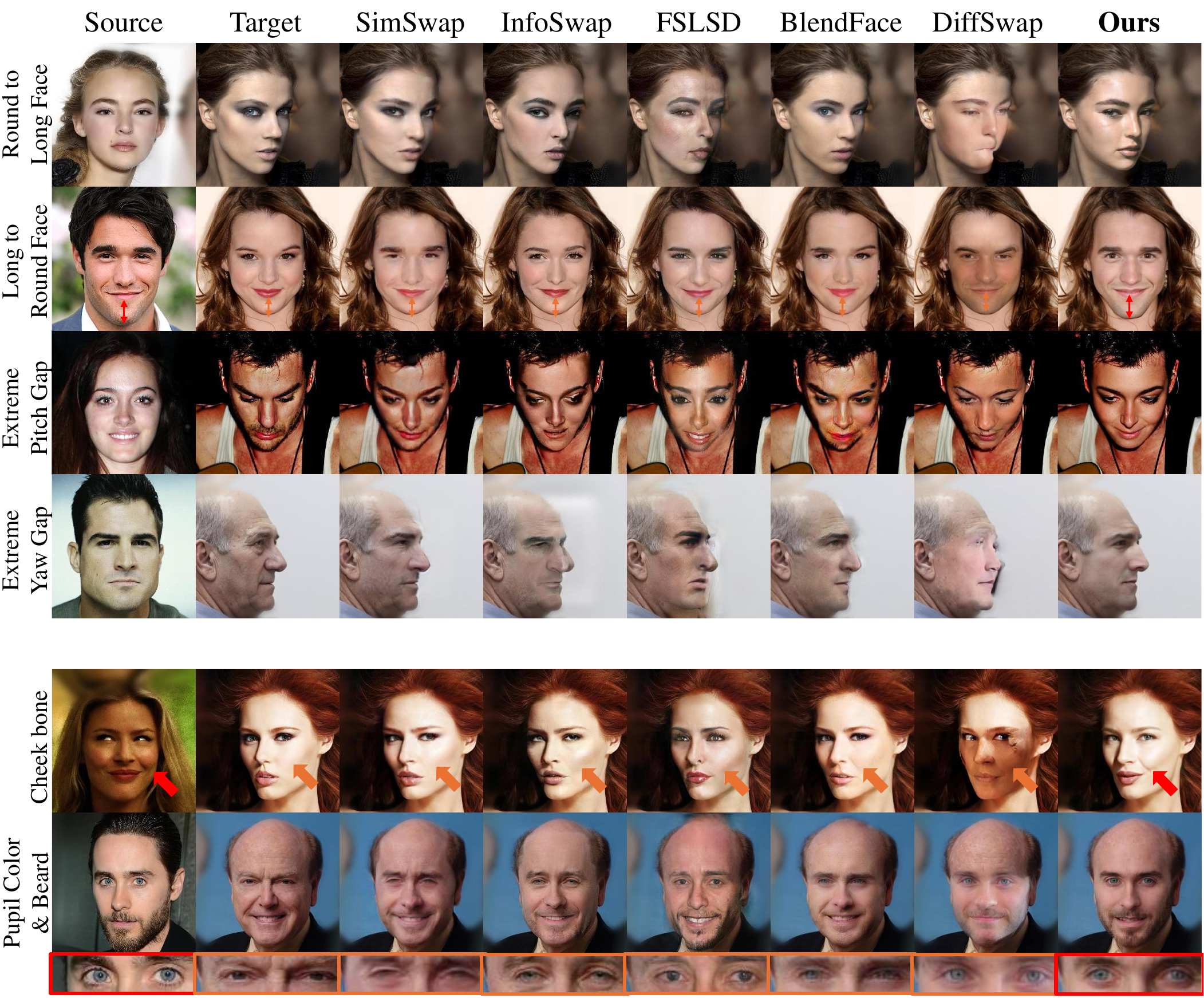}

   \caption{\textbf{Comparison among target-oriented baselines}. (Top) Other baselines struggle to replicate the source's facial features such as facial contours and volumes (\eg, jaw shape and facial scale) and (Bottom) inner facial traits (\eg, pupil color, beard, and cheekbone). In contrast, \textbf{Ours} conveys these with high-fidelity.  Pay attention to the \textcolor{red}{red} and \textcolor{orange}{orange} indicators for detailed comparison.}
   \label{fig:to_comp}
\end{figure*}

\subsection{Perforation Confusion}
In the training regime, $M$ and $I_\textit{ras}$ can both carry the same information of the facial contour of $I$.
This means that the model can attend to either $M$ or $I_\textit{ras}$ to reconstruct the facial contour of $I$.
However, in the inference phase, the final mask $M$ is constructed from the target image $I^\textit{tgt}$, whereas $I_\textit{ras}^\textit{swap}$ contains the facial contour of the source image.
Since facial contour is part of the identity (\textcolor{olive}{C2}), $I_\textit{ras}$ (or $I_\textit{ras}^\textit{swap}$ in the inference phase)  should be the sole factor for carrying the contour information.

To address this problem, we introduce a novel technique called \textit{perforation confusion} which randomly augments the final mask $M$ so that the model learns not to retrieve any contour related information from the masked image $I_\textit{p}$ (or $I_\textit{p}^\textit{tgt}$ in the inference phase). 
Specifically, the perforation confusion is performed by the following sequence:
\begin{enumerate}

    \item Sample a random shape parameter $\alpha^\textit{rand}$ from a facial image dataset~\cite{stylegan2}.
    \item Replace $\alpha$ of $v$ to $\alpha^\textit{rand}$ to construct a random 3DMM parameter tuple $v^\textit{rand}= (\alpha^\textit{rand}, \beta, \gamma_\textit{neu}, \delta, R, t_{xy}, s)$, and extract a random mask $M^\textit{rand}_\textit{ras}$ from $I^\textit{rand}_\textit{ras} = Rd(v^\textit{rand})$.
    \item $M_\textit{ras}$ is summated with $M^\textit{rand}_\textit{ras}$ and exclude non-facial region $M_\textit{occ}$ to generate the final mask $M = M_\textit{ras} + M^\textit{rand}_\textit{ras} - M_\textit{occ}$.
\end{enumerate}

\vspace{0.1cm}
\noindent During the inference phase, it's important to highlight that we substitute $M^\textit{rand}_\textit{ras}$ with $M^\textit{swap}_\textit{ras}$ which is directly derived from the foreground region of $I^\textit{swap}_\textit{ras}$, and the final mask is generated as $M = M_\textit{ras}^\textit{tgt} + M_\textit{ras}^\textit{swap} - M_\textit{occ}$.
Note that $M_\textit{ras}^\textit{tgt}$ is a foreground mesh mask derived from the target facial mesh $I^\textit{tgt}_\textit{ras}$ of the target image $I^\textit{tgt}$. 
The necessity of the perforation confusion technique for handling the shape-misalignment problem of source and target face is demonstrated by the results in Fig.~\ref{fig:concept} (B) and Sec.~\ref{subsec:ablation}.

\begin{table}[t!]
    \caption{\textbf{Quantitative comparison} with both target-oriented and source-oriented baselines. \textbf{Bold} text highlights the best scores. Our method outperforms other baselines in Identity Similarity (ID. Sim.), Identity Consistency (ID. Cons.), Head Pose, and Fréchet Inception Distance (FID). Expression distance is on par with the best performing models.}
    \centering
    \begin{adjustbox}{width=1\textwidth}

    \small
    \begin{tabular}{lccccccc}
    \Xhline{2.5\arrayrulewidth}
    Categories & Methods &  ID. Sim.$\uparrow$ &  ID. Cons.$\uparrow$ & Shape $\downarrow$ & Expression$\downarrow$ & Head Pose$\downarrow$ & FID$\downarrow$ \\
    \midrule
    \multirow{4}{*}{\textbf{Target-Oriented}} 
    & SimSwap~\cite{simswap}    & 0.525 & 0.543 &0.128 &0.204 & 0.014 & 26.77 \\
    & InfoSwap~\cite{infoswap}  & 0.527 & 0.583 &0.126& 0.233 & 0.019 & 32.21 \\
    & FSLSD~\cite{fslsd}        & 0.330 & 0.345 &0.129& 0.207 & 0.025 & 39.71 \\
    & BlendFace~\cite{blendface} & 0.440 & 0.510 &0.136& \textbf{0.189} & \textbf{0.013} & 23.11 \\
    & DiffSwap~\cite{blendface} & 0.347 & 0.361 &0.156 & 0.224 & 0.028 & 59.98 \\
    \midrule
    \multirow{2}{*}{\textbf{Source-Oriented}} 
     &{FSGAN~\cite{fsgan}}   &0.338&0.403& 0.164&0.193&0.016 & 42.56 \\ 
     & {E4S~\cite{e4s}}     &0.501 &0.588& 0.118&0.262&0.028 & 52.90\\
    \midrule

    \textbf{SAMAE} & {\textbf{Ours}} & \textbf{0.578} & \textbf{0.628} & \textbf{0.108}&0.190 & \textbf{0.013} &\textbf{21.22}\\
    \Xhline{2.5\arrayrulewidth}
    \end{tabular}
    \end{adjustbox}
    \label{table:baselines_to}
\end{table}

\noindent\subsection{Random Mesh Scaling}
Empirically, during the inference phase, we observed that when the facial volume of the source is significantly smaller (or larger) than that of the target, the swapped face appears awkward, either shrunken or dilated. 
This issue arises because the model, trained to self-reconstruct the input image with a consistently sized facial region, tends to over-rely on the pixel-aligned information from $I_\textit{ras}$.
This reliance hinders generalization to cross-identity inferences.

To handle this problem, we propose the \textit{random mesh scaling} technique, which allows the model to generate realistic face images using randomly scaled $I_\textit{ras}$, thereby enhancing the model's ability to generalize to facial priors of varying scales during inference.
Specifically, in the train regime, the scale parameter $s$ is substituted with a random scale parameter $s^\textit{rand} \sim U(-4,1)$, leading to randomly scaled $I_\textit{ras}$.
It is important to note that random mesh scaling is not used during the inference phase; instead, the estimated target scale $s^\textit{tgt}$ is utilized.

\subsection{Disentangling Albedo Condition} 
\label{subsec:skin_color} 
The estimated 3DMM albedo parameter $\gamma$ contains both non-identity attributes such as skin color (\textcolor{olive}{C1}), and identity features, including skin details and certain inner facial traits like eye color (\textcolor{olive}{C2}). 
Due to the entangled nature of these attributes within $\gamma$, it is unsuitable for our task that requires distinct separation of identity and non-identity features.
Given the necessity of the albedo parameter in rendering $I_\textit{ras}$, we have devised a workaround. We transform $\gamma$ into its neutralized form, $\gamma_\textit{neu}$, which incorporates a white-colored albedo map.
This modification eliminates any albedo-related information from $I_\textit{ras}$. Furthermore, $I_\textit{ras}$ undergoes min-max normalization to isolate the remaining brightness differences, which serve as illumination information in a facial image.

Instead of relying on albedo parameter $\gamma$, we find that using identity embeddings $c_\textit{id}$ extracted from $E_\textit{FR}$ is sufficient, as they contain rich identity information.
However, directly using these can lead to unintended transfer of skin color from the source face, which should be derived from the target face.
To address this, we apply random color-jittering to the image before processing it with $E_\textit{FR}$, encouraging our generator to disregard skin color information in $c_\textit{id}$.

In parallel, we train an additional trainable encoder $E_\textit{skin}$ to capture skin color information as a vector-formed neural albedo.
Specifically, we mask out non-skin areas from $I$ (or $I^\textit{tgt}$ during inference) with an off-the-shelf face parser~\cite{bisenet}, and encode the masked image with $E_\textit{skin}$.
To prevent $c_\textit{skin}$ from incorporating irrelevant features such as skin detail or facial geometry, we empirically choose a low-dimensional embedding, ensuring it has only the necessary capacity to capture skin color information.
More details and experiments are provided in the supplementary materials.

\subsection{Training Objectives} 
We employ reconstruction losses (L1 loss and Perceptual loss~\cite{vgg}) and an identity loss~\cite{arcface} between the estimated image $\hat{I}$ and the ground-truth image $I$. 
To enhance the realism of the output image, we also incorporate a non-saturating adversarial loss~\cite{stylegan2}.
For additional details, please refer to the supplementary materials.

\section{Experiments}
\textbf{Datasets.} 
The model is trained on FFHQ dataset~\cite{stylegan2} with images at a resolution of $256\times256$.
For evaluation, we use 1K randomly sampled source-target pairs, from the CelebA-HQ dataset~\cite{sg1}.
It is noteworthy that, unlike our method, numerous approaches~\cite{styleswap,faceshifter,reinforceswap} extend their training datasets beyond FFHQ by integrating multiple datasets, including identity-labeled datasets and/or video datasets~\cite{vggface2,voxceleb} to stabilize the training.
Our method, on the other hand, delivers state-of-the-art performance without such additions.

\noindent\textbf{Baselines.} For both quantitative and qualitative comparisons, we establish our baselines with FSGAN~\cite{fsgan}, SimSwap~\cite{simswap}, InfoSwap~\cite{infoswap}, FSLSD~\cite{fslsd}, E4S~\cite{e4s}, DiffSwap~\cite{diffswap}, and BlendFace~\cite{blendface}.
We exclude HifiFace~\cite{hififace} and FaceShifter~\cite{faceshifter} from our main analysis, as they lack official open-sourced codes. 
Additionally, in the supplementary materials, we present comparisons with other methods such as AFS~\cite{afs}, MegaFS~\cite{megafs}, and ReliableSwap~\cite{reliableswap}, which either demonstrate subpar quality or have not been accepted to the academic community.

\noindent\textbf{Implementation Details.} 
Our model is trained on two NVIDIA RTX 3090 GPUs with a batch size of 8, completing 500k iterations in about 6 days.
We use the Adam optimizer with a learning rate of $2\times10^{-4}$ for both the generator and the discriminator.
We leverage the ADM~\cite{adm} U-Net for the generator, and StyleGAN2~\cite{stylegan2} architecture for the discriminator.

\subsection{Qualitative Comparisons}
We categorize the baselines into two groups: target-oriented approaches including SimSwap~\cite{simswap}, InfoSwap~\cite{infoswap}, FSLSD~\cite{fslsd}, BlendFace~\cite{blendface}, and DiffSwap~\cite{diffswap} and source-oriented approaches FSGAN~\cite{fsgan} and E4S~\cite{e4s}.

\noindent\textbf{Target-oriented Baselines.} Target-oriented baselines directly employ reconstruction losses between generated outputs and target images, effectively preserving target attributes like lighting conditions, expressions, and head poses, as shown in Fig.~\ref{fig:to_comp}.
However, this straightforward application of reconstruction loss often leads to the leakage of the target's identity, resulting in a blend of source and target identities. 
These models frequently struggle to accurately represent the source's facial contours and skin details. 
In contrast, our model moves beyond this trade-off, adeptly reproducing the source's facial contours, skin details, and inner facial features, while still maintaining the target's non-facial attributes, and poses. 
Further results can be found in the supplementary materials.

\noindent\textbf{Source-oriented Baselines.} 
We present a comparison of our method with source-oriented baselines in Fig.~\ref{fig:so_comp}. 
These methods involve merging the reenacted source face with the target image, effectively reducing the risk of target identity leakage.
However, their effectiveness is limited by the reenactment models' performance, often leading to inaccuracies in source identity preservation and difficulties in handling pose variations.
Furthermore, the blending process can result in unnatural skin tones and inadequate replication of the target's lighting, sometimes introducing noticeable artifacts.
For example, the shadow from the source image can be carried over into the swapped image, as depicted in the second row of Fig.~\ref{fig:so_comp}. 
In contrast, our method is adept at matching the target's skin color and lighting conditions, producing outputs that blend seamlessly with the target images.



\definecolor{brightube}{rgb}{0.863, 0.784, 0.871}

\begin{figure}[tb]
  \centering
  \begin{subfigure}{0.53\linewidth}
    \includegraphics[width=\linewidth]{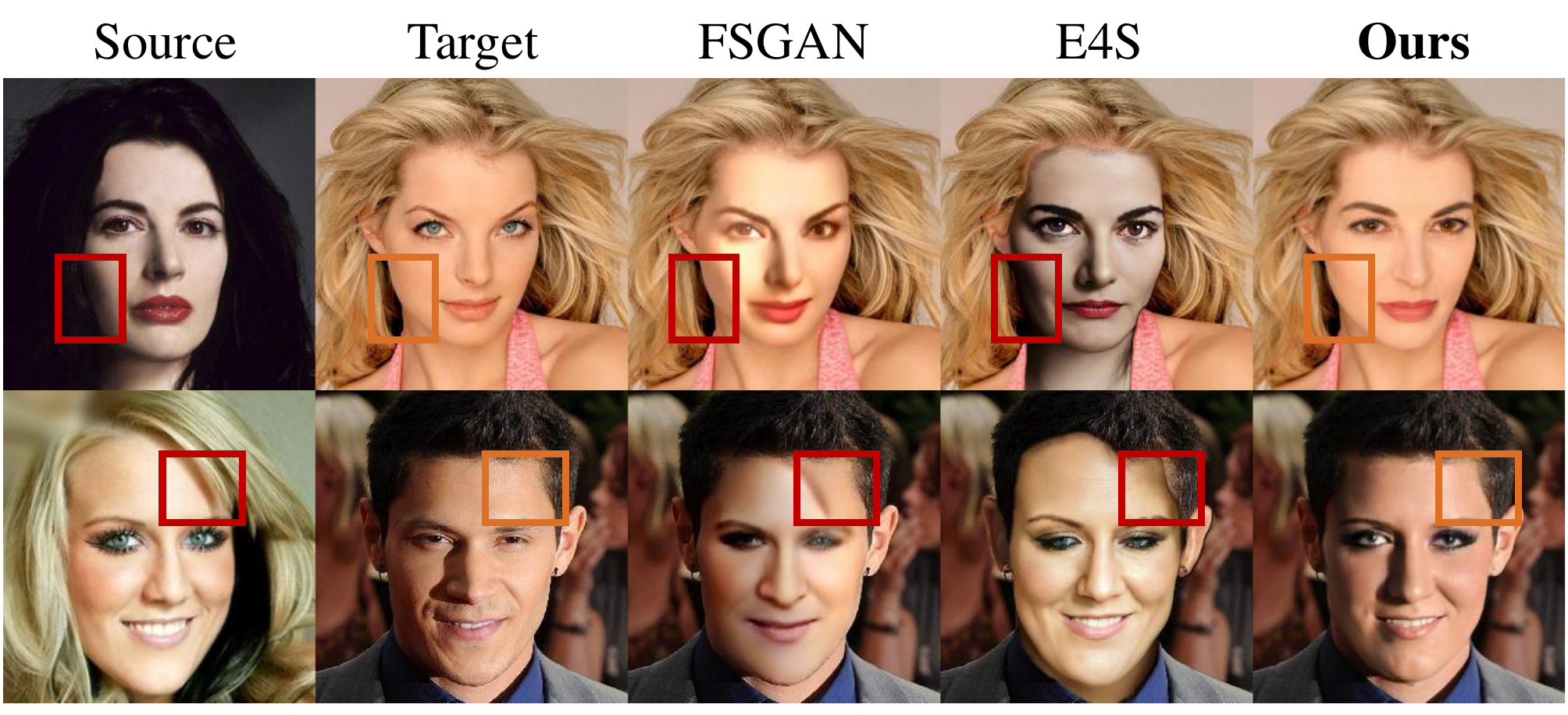}
    \caption{\textbf{Comparison among source-oriented baselines.} These baselines exhibit issues with leakage of the source's illumination. Observe \textcolor{red}{red} and \textcolor{orange}{orange} indicators. Our model effectively avoids source illumination leakage, thanks to our method's finely disentangled features.}
    \label{fig:so_comp}
  \end{subfigure}
  \hfill
  \begin{subfigure}{0.46\linewidth}
   \includegraphics[width=1.03\linewidth]{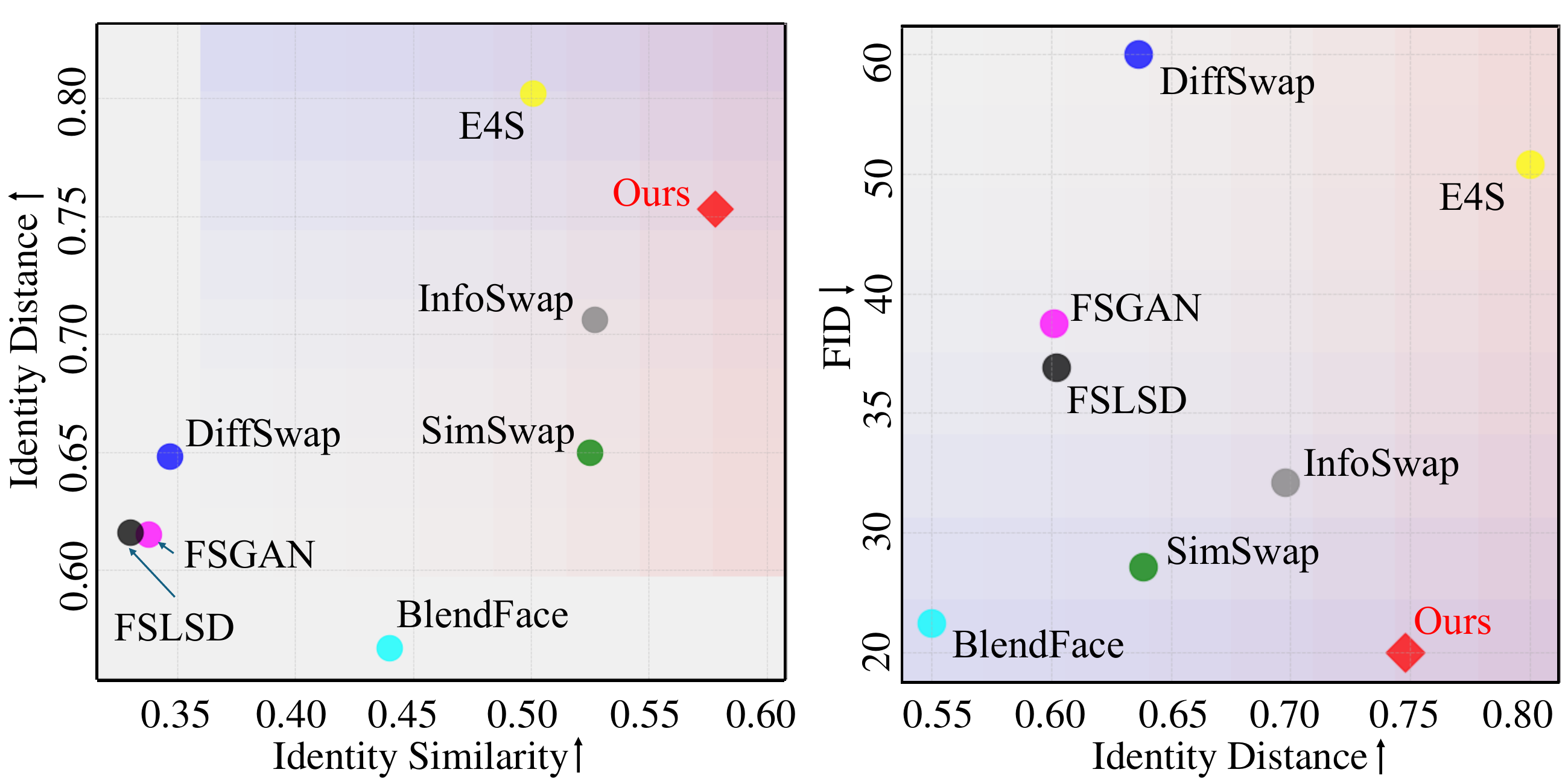}
    \caption{2-dimensional graph comparison on Identity Similarity and Distance (left), where positioning in the \textcolor{brightube}{upper-right} corner signifies a good model. In the Identity Distance - FID graph (right), a location in the \textcolor{brightube}{lower-right} side indicates favorable model performance.}
    \label{fig:id_sim_dis}
  \end{subfigure}
  \caption{Comparison among source-oriented baselines and 2-dimensional graph comparison on Identity scores and FID.}
  \label{fig:short}
\end{figure}

\subsection{Quantitative Comparisons}
We conduct a quantitative comparison of our method against both target-oriented and source-oriented approaches. 
The evaluation involves several metrics: Identity Similarity (ID. Sim.), Identity Consistency (ID. Cons.), Shape, Expression and Head Pose distance, and the Fréchet Inception Distance (FID) score. 
Identity Similarity (ID. Sim.) calculates the cosine similarity between the identity embeddings of the source and swapped images, using a separate face recognition model~\cite{arcface} that was not used in the training phase. 
Identity Consistency (ID. Cons.)~\cite{reinforceswap} measures the consistency of identity among the swapped images, using one source image and various target images. 
Further, we measure the Expression and Head Pose Distance between the target and swapped faces, as well as the Shape Distance between the source and the swapped results, using a 3DMM predictor~\cite{deca} that was also excluded from the training process.
The FID score evaluates the overall quality of the images.

As shown in Table~\ref{table:baselines_to}, our method achieves the state-of-the-art quality without sacrificing certain metrics, unlike other methods.
InfoSwap ranks second-best in terms of ID. Sim. but falls behind in other metrics.
While DiffSwap tackles the shape misalignment problem, the Shape Distance results suggest shortcomings in the method.
Regarding Expression distance, although BlendFace achieves the highest score, it exhibits a lower ID. Sim. score. 
Additionally, its low ID. Cons. score suggests a leakage of target attributes. This indicates that BlendFace potentially compromises identity preservation in favor of maintaining the target's pose and expression.

Also to comprehensively evaluate our model's performance, including aspects such as target identity leakage and image realism, we present a two-dimensional graph in Fig.~\ref{fig:id_sim_dis}.
In Fig.~\ref{fig:id_sim_dis} (left), the $x$-axis shows the identity similarity score between the source and the generated images, while the $y$-axis indicates the identity distance between the target and the generated images.
Optimal models without target identity leakage should exhibit both high identity similarity scores and high identity distances, positioning them in the upper-right corner of the graph.
The figure implies that our method excels in reflecting the source identity and preventing target identity leakage. 
E4S is also positioned at the upper-right corner of the figure.
However, right panel of Fig.~\ref{fig:id_sim_dis} indicates that E4S yields the lowest FID score, suggesting it creates source-like images with reduced target identity leakage, but with poor realism. 
In contrast, our model is situated in the bottom-right corner of the graph, demonstrating superior performance in image realism and robustness against target identity leakage, outperforming other baselines.

\subsection{Ablation Study}
\label{subsec:ablation} 
\begin{figure}[tb]
  \centering
  \begin{subfigure}[t]{0.59\linewidth}
    \includegraphics[width=\linewidth]{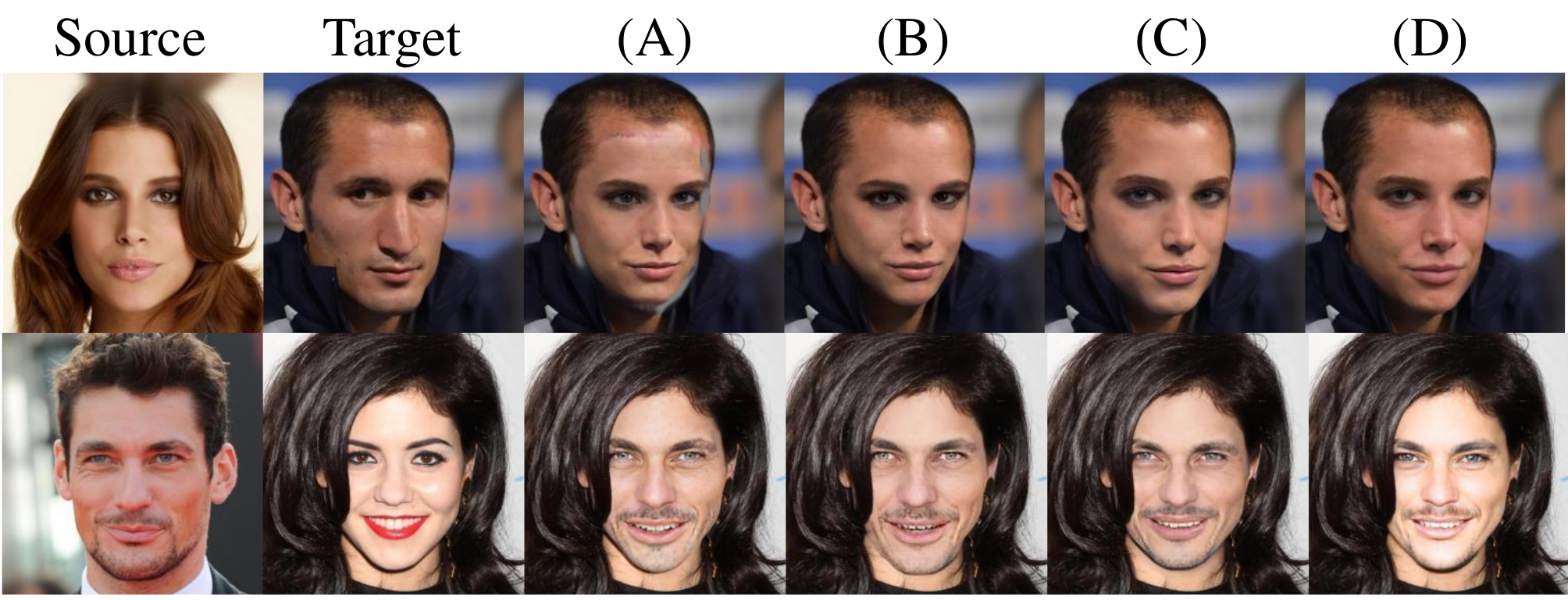}
    \caption{\textbf{Qualtitative evaluation in ablation study.} (A) Ours Base, (B) Ours Base + perforation confusion (\color{cyan}{P}\color{black}{)}, (C): Ours Base + perforation confusion (\color{cyan}{P}\color{black}{)} + random mesh scaling (\color{lime}{R}\color{black}{)}, and (D): Ours Base + perforation confusion (\color{cyan}{P}\color{black}{)} + random mesh scaling (\color{lime}{R}\color{black}{)} + skin color condition (\color{orange}{S}\color{black}{)}.}
    \label{fig:short-a}
  \end{subfigure}
  \hfill
  \begin{subfigure}[t]{0.4\linewidth}
    \includegraphics[width=\linewidth]{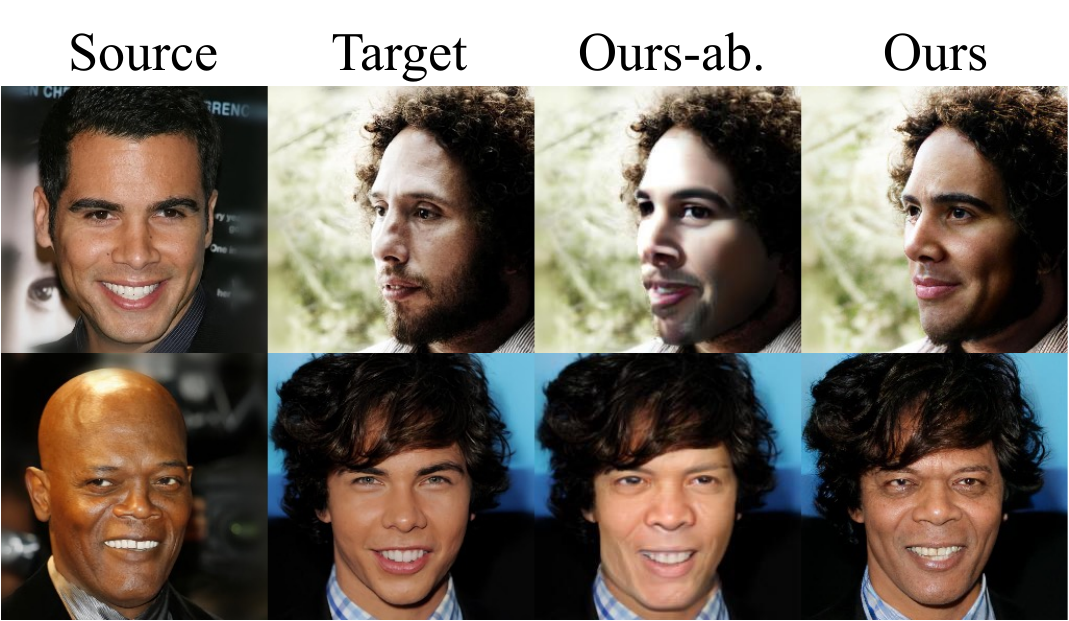}
    \caption{Qualtitative comparison between the seesaw game training scheme (Ours-ab.) and the self-supervised training scheme (Ours).}
    \label{fig:short-b}
  \end{subfigure}
  \caption{Qualitative results for various ablation studies.}
  \label{fig:ablation_holistic}
\end{figure}

We conducted an ablation study on various components of our method, including perforation confusion, random mesh scaling, and disentangled albedo conditions. 
The results are shown in Fig.~\ref{fig:short-a}.
Column (A) displays the base model without these techniques.
Without perforation confusion, the model struggles with cross-identity swapping and fails to properly inpaint the gaps caused by shape misalignment between the source and target faces.
With perforation confusion, as shown in column (B), the model handles the shape misalignment and successfully fills non-facial areas, such as the neck. 
However, this column also reveals that without random mesh scaling, the facial volume in the target image is not preserved, leading to swapped images with unnaturally shrunken or enlarged faces.
Random mesh scaling allows the model to adaptively fit the target priors, maintaining the facial volume of the target image (column (C)).
Finally, matching the skin color of the generated results with the target images is crucial for seamless blending. 
By employing the disentangled skin color embedding $c_\textit{skin}$, we see in column (D) that the skin color and illumination of the swapped faces closely match those of the target, enhancing the realism of the blended images.

Also to assess the effectiveness of SAMAE's self-supervised training approach compared to the widely used seesaw game training regime, we conducted a comparison against a model trained using a multi-task learning strategy similar to target-oriented methods.  
Specifically, the model generates a swapped image from randomly selected source and target images, and both the identity loss and reconstruction loss are applied to the output.
We utilized the same network architecture and hyperparameters as our original model for this experiment.
The resulting output generally exhibits blurriness and incorrect identity reflectance and illumination, as demonstrated in Fig.~\ref{fig:short-b}, labeled as Ours-ab. 
We also provide quantitative evaluation results in the supplementary materials.

\begin{table}[t]
    \caption{\textbf{Quantitative evaluation in ablation study.} (B) indicates Ours Base. Perforation confusion (\color{cyan}{P}\color{black}{)} enhances realism (FID), reducing unpainted regions or cut-off facial contours. Random mesh scaling (\color{lime}{R}\color{black}{)} further boosts the realism and identity scores. Incorporating skin color condition (\color{orange}{S}\color{black}{)} ensures skin color and illumination closely match the target images, yielding naturally blended images, and optimally improves all metrics.}
    \centering
    \small
    \begin{tabular}{lcccc}
    \Xhline{2.5\arrayrulewidth}
    Methods &  ID. Sim. $\uparrow$  & Expression $\downarrow$ & Head Pose $\downarrow$  & FID $\downarrow$ \\
    \Xhline{.8\arrayrulewidth}
   
B  &0.570 &0.208&0.014& 24.05\\ 
B+\color{cyan}{P} &0.568 & 0.209 &0.014& 23.92\\
B+\color{cyan}{P}\color{black}{+}\color{lime}{R} &0.573 & 0.220 &0.015& 22.99\\
B+\color{cyan}{P}\color{black}{+}\color{lime}{R}\color{black}{+}\color{orange}{S} &\textbf{0.578} &\textbf{0.190} &\textbf{0.013}   &\textbf{21.22}\\

    \Xhline{2.5\arrayrulewidth}
    \end{tabular}
    
\label{table:ablation}
    
\end{table}

%

\begin{figure}[tb]
  \centering
  \begin{subfigure}[t]{0.49\linewidth}
   \includegraphics[width=\linewidth]{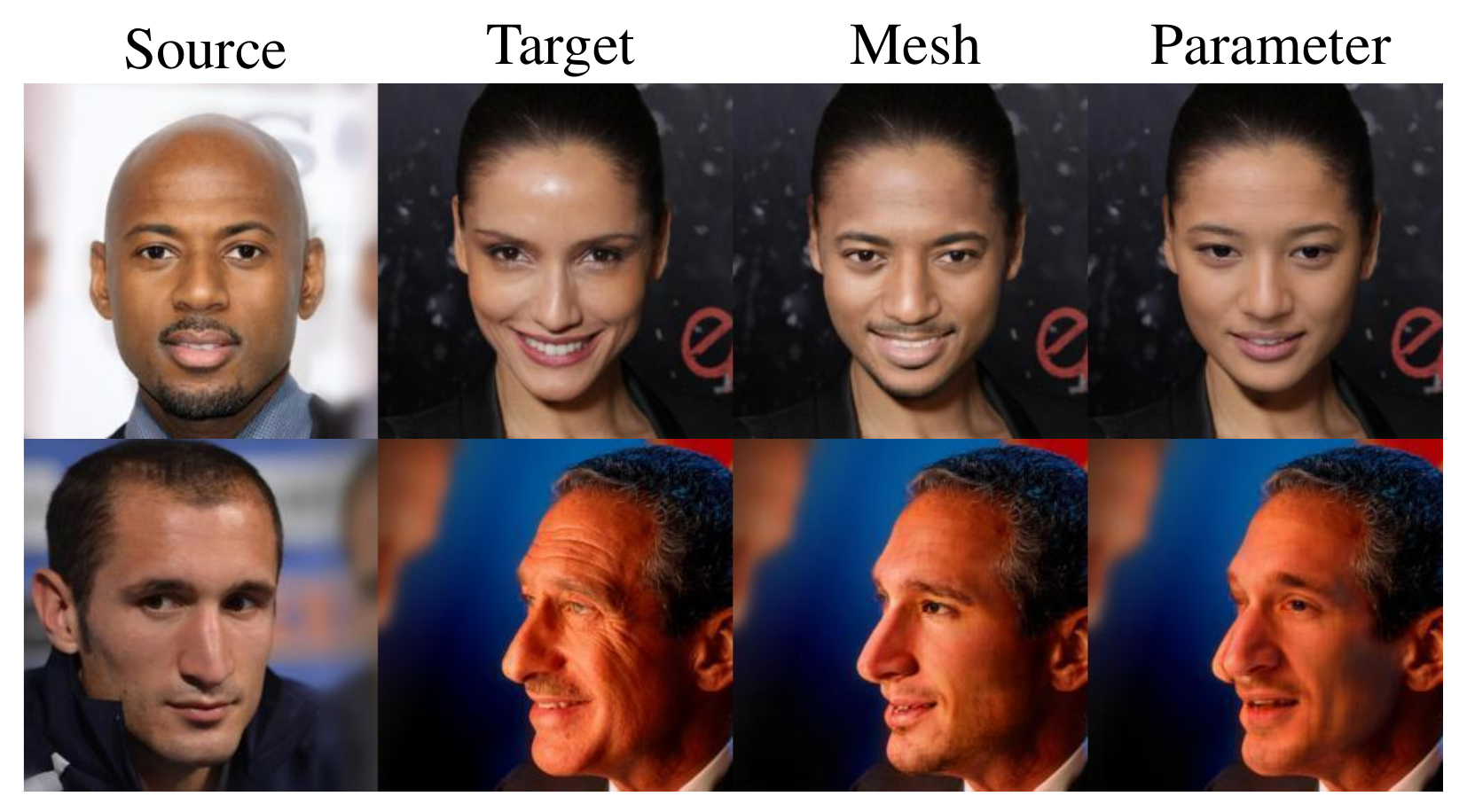}
    \caption{\textbf{Another option on the geometry condition.} Models trained with direct 3DMM parameters in vector form, without rendering (Parameter), often fail to accurately capture the precise facial geometries of the source.}
   \label{fig:mesh_vs_param}
  \end{subfigure}
  \hfill
  \begin{subfigure}[t]{0.49\linewidth}
   \includegraphics[width=\linewidth]{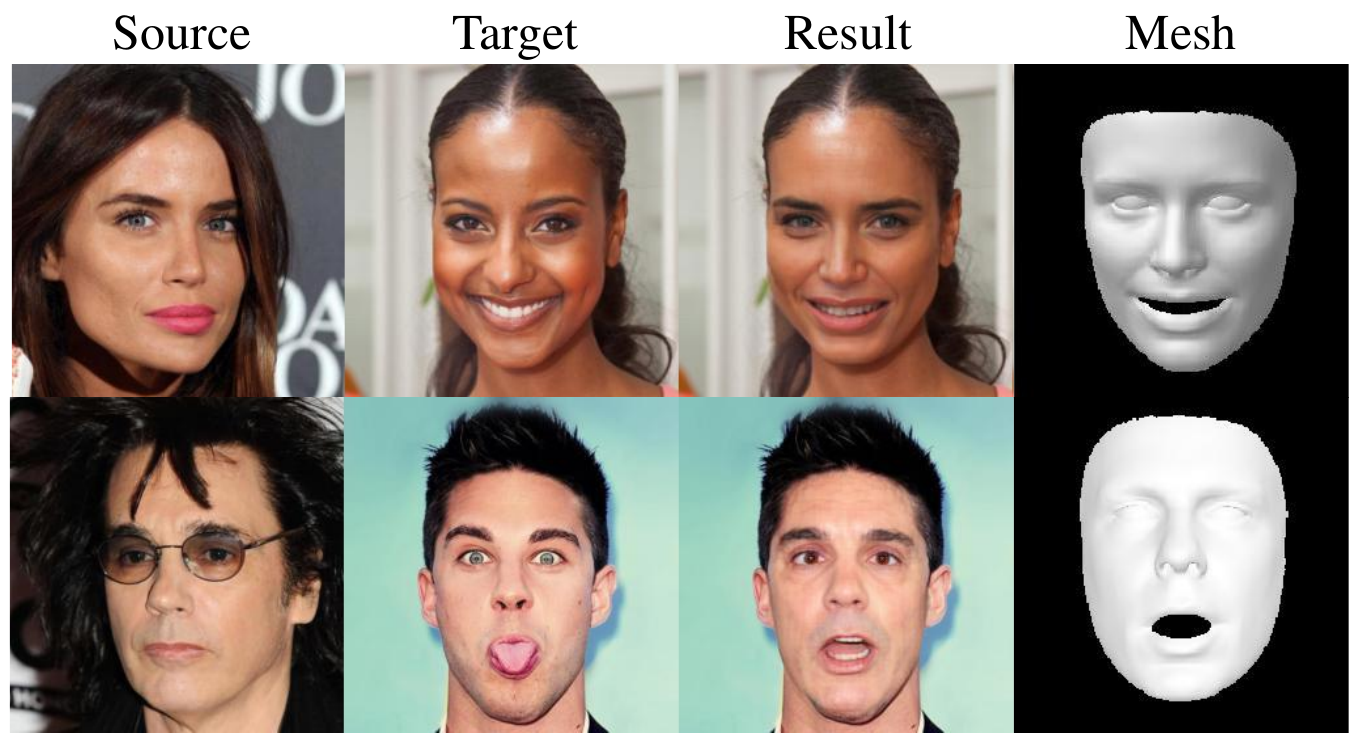}
    \caption{\textbf{Limitations}. When 3DMM estimators are unable to accurately capture exaggerated expressions or tongue, as shown in the ``Mesh'' visualization, our model is constrained by the mesh condition, reflecting these limitations.}
    \label{fig:limitation}
  \end{subfigure}
\caption{Selection of geometry conditions and limitations of our model.}

\end{figure}

We also compare two different conditioning methods: ``parameter'' and ``mesh''. 
The mesh condition, as described in our paper, uses $I^\textit{swap}_\textit{ras}$ for the geometry condition, fully utilizing 3DMM renderer and thus providing the model with enhanced spatial information.
The parameter method directly employs 3DMM coefficients $v^\textit{swap}$ in vector form without rendering the mesh.
As shown in the first row of Fig.~\ref{fig:mesh_vs_param}, we find that the mesh condition is vital for preserving the source's facial features.
Furthermore, the results in the second row indicate that the mesh condition more accurately guides the model in learning the head pose. 
On the other hand, the parameter condition results in an artifact on the nose with messy geometry.





\section{Discussion}
\textbf{Limitations and Future Work.} As can be seen in Fig.~\ref{fig:limitation}, our model sometimes fails to capture detailed expressions, such as exaggerated grin, due to the limited representation capacity of 3DMM. 
Advanced 3DMMs and their prediction models~\cite{deca,flame} can be utilized for future works.
We posit that incorporating strong generative priors like StyleGAN or the latest diffusion models into our training pipeline could enhance the face swapping quality and will be an interesting research direction.

\noindent\textbf{Ethical Considerations.} 
Face swapping is useful in areas like digital resurrection and telepresence but also poses risks of privacy invasion and misinformation. 
We are dedicated to prevent the potential misuse our model, and plan to release our model exclusively for research purposes. Additionally, we will provide a benchmark dataset to support research in face forensics and privacy protection.

\clearpage

\noindent\textbf{Acknowledgements.} This work was supported by Institute for Information \& communications Technology Promotion(IITP) grant funded by the Korea government(MSIT) (No.RS-2019-II190075 Artificial Intelligence Graduate School Program(KAIST)), the National Research Foundation of Korea (NRF) grant funded by the Korea government (MSIT) (No. NRF-2022R1A2B5B02001913), and KAIST-NAVER hypercreative AI center.


%
%
\bibliographystyle{splncs04}
\bibliography{main}
\end{document}


\pagestyle{plain}

\title{SelfSwapper: Self-Supervised Face Swapping via Shape Agnostic Masked AutoEncoder - Supplementary Materials}

\titlerunning{SelfSwapper}

\author{Jaeseong Lee\inst{*} \and
Junha Hyung\inst{*} \and
Sohyun Jung \and
Jaegul Choo 
}
\authorrunning{J. Lee et al.}

\institute{KAIST AI
\\
\email{\{wintermad1245, sharpeeee, jsh0212, jchoo\}@kaist.ac.kr 
} \\
* indicates equal contributions.}
\maketitle
\section{CelebA-HQ Comparison with AFS, MegaFS, and ReliableSwap}

In the Experiment section of our manuscript, we conduct a comprehensive comparison with additional baselines, including MegaFS~\cite{megafs}, AFS~\cite{afs}, and ReliableSwap~\cite{reliableswap}. These baselines, although not widely accepted within the academic community, provide a valuable contrast in terms of their performance and capabilities. The quantitative results are presented in Table~\ref{supp_table:baselines_more}, extending the comparison beyond Table 1 in the main manuscript. Additionally, we introduce an overall score by standardized for each metric and averaged following the methodology of SmoothSwap~\cite{smoothswap}, denoted \textit{Overall}. Among the nine baselines, our approach achieves the highest overall score, indicating a substantial performance gap.

\begin{figure*}[t]
  \centering
   \includegraphics[width=0.96\linewidth]{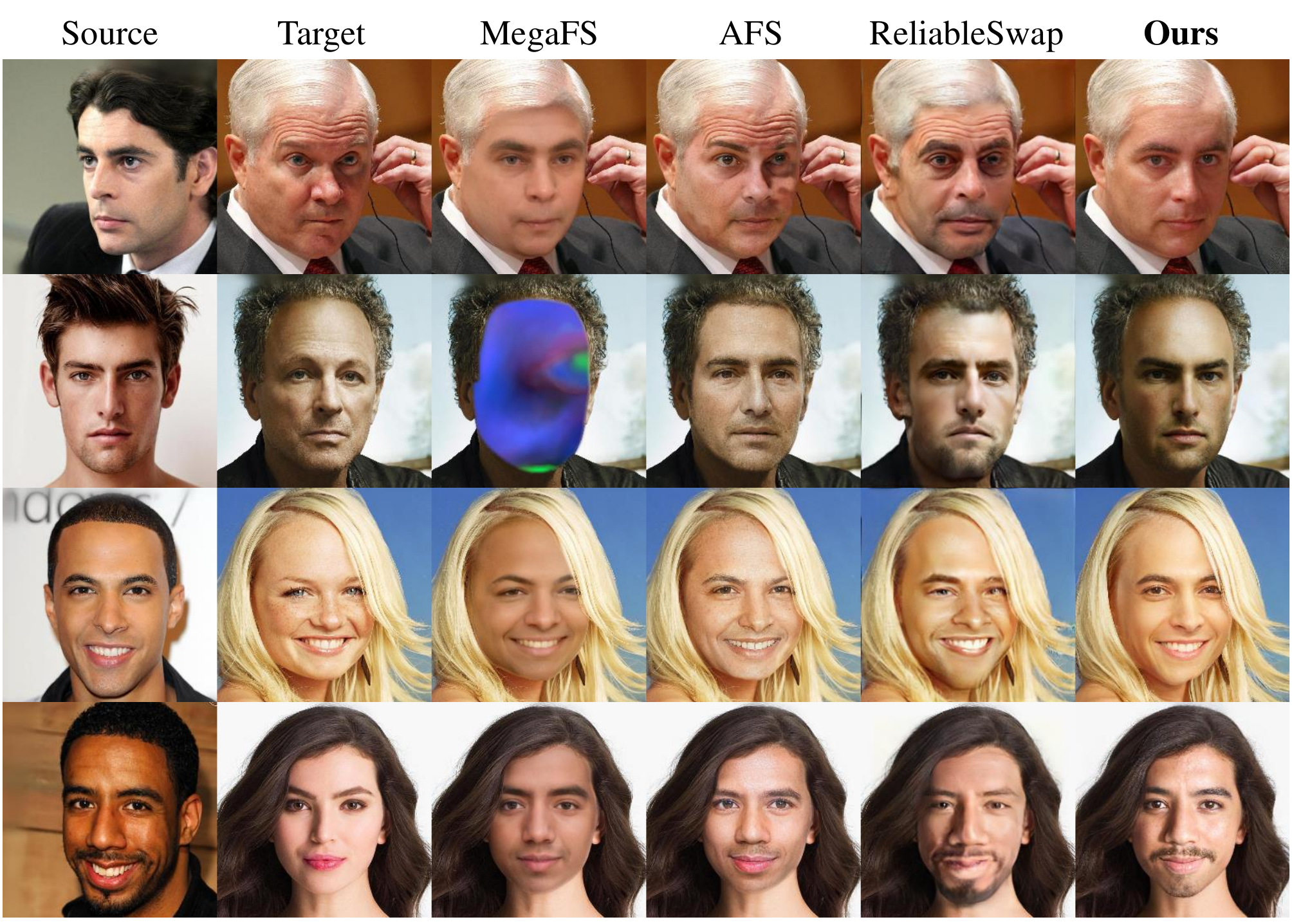}

   \caption{Qualitative comparison of additional baselines (MegaFS~\cite{megafs}, AFS~\cite{afs}, and ReliableSwap~\cite{reliableswap}).}
   \vspace{-0.2cm}
   \label{fig:additional_baseline1}
\end{figure*}

\begin{figure*}[t]
  \centering
   \includegraphics[width=0.96\linewidth]{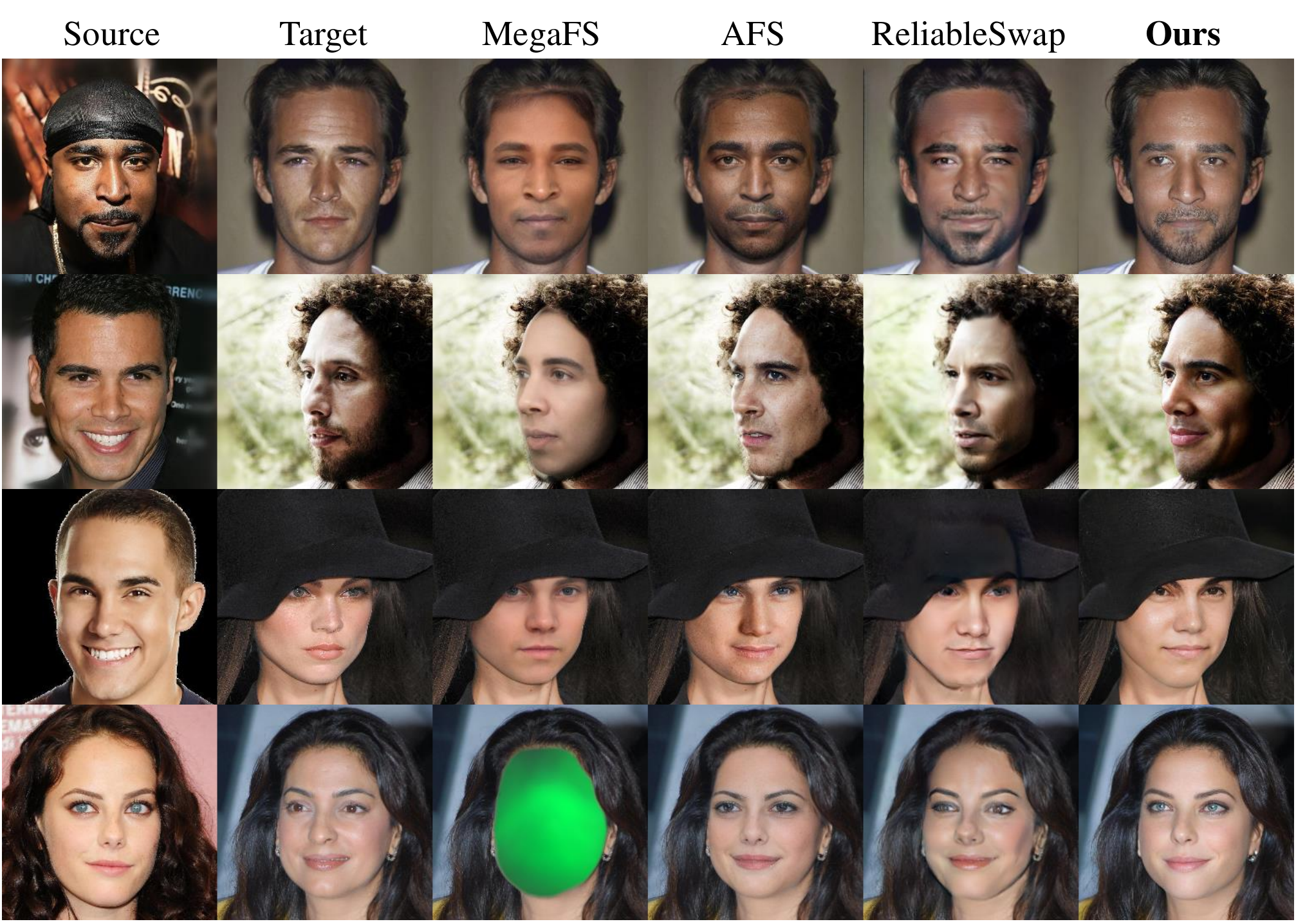}

   \caption{Qualitative comparison of additional baselines (MegaFS~\cite{megafs}, AFS~\cite{afs}, and ReliableSwap~\cite{reliableswap}).}
   \label{fig:additional_baseline2}
\end{figure*}

A closer examination of MegaFS and AFS, depicted in the 3rd and 4th columns of Fig.~\ref{fig:additional_baseline1} and~\ref{fig:additional_baseline2}, reveals limitations in handling target attributes such as illumination and skin color. These methods rely on an encoder-based StyleGAN~\cite{stylegan2} inversion with a empirical layer-specific decomposition design. Moreover, they necessitate post-processing to address issues like hair and background, as StyleGAN inversion-based techniques often freeze and map latent vectors, leading to insufficient attribute preservation.

ReliableSwap, illustrated in the 5th columns of Fig.~\ref{fig:additional_baseline1} and~
\ref{fig:additional_baseline2}, exhibits unrealistic results with noticeable artifacts. While attempting to introduce reliable supervision through pseudo-triplets for face swapping training, ReliableSwap depends on the head reenactment model~\cite{lia} to construct the synthetic dataset. This reliance on the head reenactment model introduces limitations, constraining the capacity of the trained model.  

\section{More Qualitative Comparison among Target-Oriented Baselines}
\begin{figure*}
  \centering
   \includegraphics[width=1\linewidth]{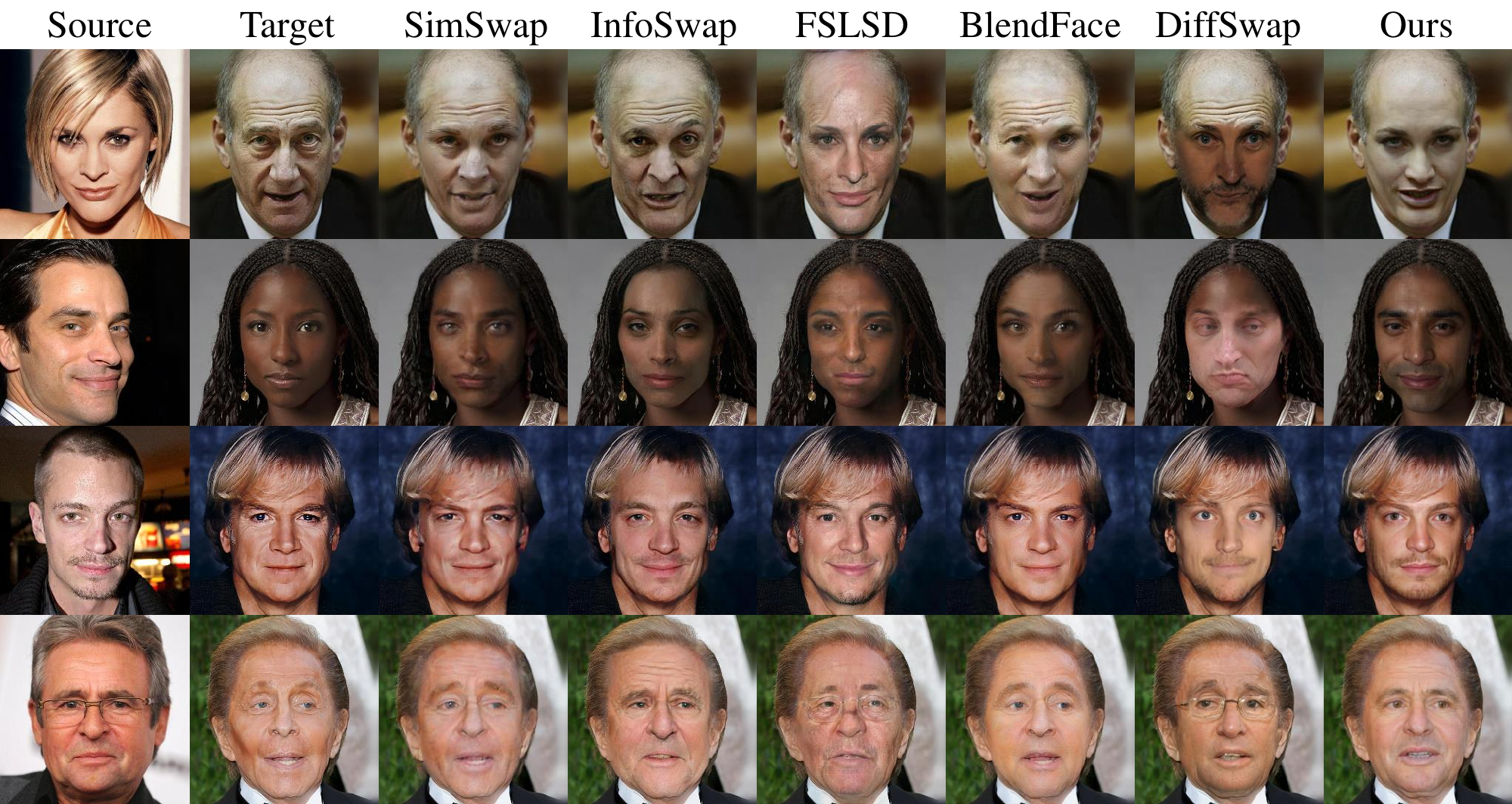}

   \caption{Additional qualitative comparison of target-oriented baselines.}
       \vspace{-1.cm}

   \label{supp_fig:to1}
\end{figure*}

\begin{figure*}
  \centering
   \includegraphics[width=1\linewidth]{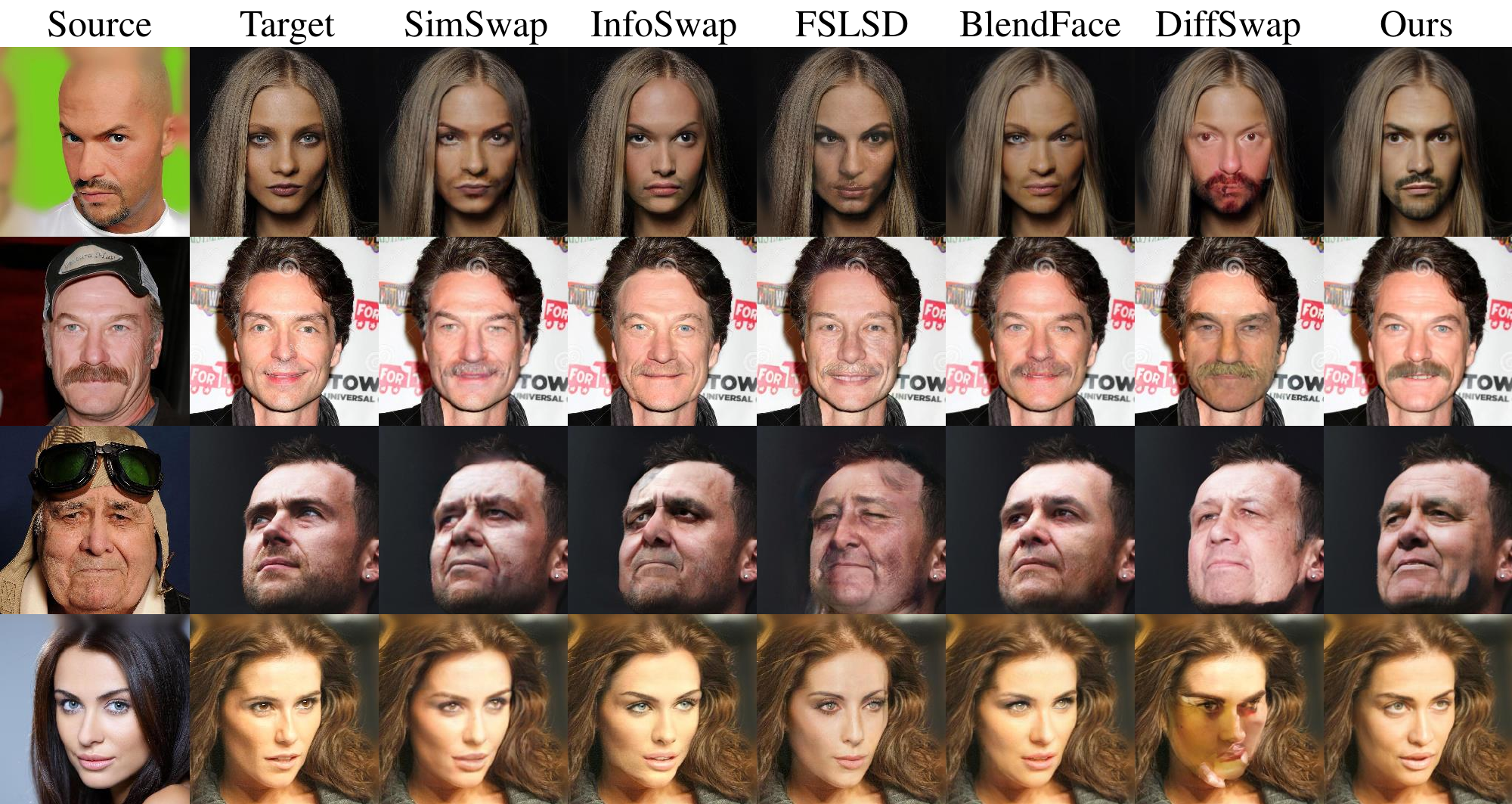}

   \caption{Additional qualitative comparison of target-oriented baselines.}
   \label{supp_fig:to2}
\end{figure*}

We have posted eight sets of additional qualitative comparison among target-oriented baselines in Fig.~\ref{supp_fig:to1} and~\ref{supp_fig:to2}.

\clearpage

\begin{table*}[h]
\caption{\textbf{Quantitative comparison augmented with MegaFS~\cite{megafs}, AFS~\cite{afs}, and ReliableSwap~\cite{reliableswap}}. Bold text highlights the best scores. The Overall score indicates average standardized score of each method.}
    \centering
    \small
    \begin{adjustbox}{width=1\linewidth}
    \begin{tabular}{lccccccc|c}
    \Xhline{2.5\arrayrulewidth}
    Categories & Methods &  ID. Sim.$\uparrow$ &  ID. Cons.$\uparrow$ & Shape $\downarrow$ & Expression$\downarrow$ & Head Pose$\downarrow$ & FID$\downarrow$ & Overall$\downarrow$ \\
    \Xhline{.8\arrayrulewidth}
    \multirow{7}{*}{\textbf{Target-Oriented}} 
    & SimSwap~\cite{simswap}    & 0.525 & 0.543 &0.128 &0.204 & 0.014 & 26.77 & -0.725 \\
    & InfoSwap~\cite{infoswap}  & 0.527 & 0.583 &0.126& 0.233 & 0.019 & 32.21 & -0.469 \\
    & FSLSD~\cite{fslsd}        & 0.330 & 0.345 &0.129& 0.207 & 0.025 & 39.71 & 0.333\\
    & BlendFace~\cite{blendface} & 0.440 & 0.510 &0.136& \textbf{0.189} & \textbf{0.013} & 23.11 &-0.593 \\
    & DiffSwap~\cite{diffswap}          & 0.347 & 0.361 &0.156 & 0.224 & 0.028 & 59.98 & 0.954 \\
    & MegaFS~\cite{simswap}             & 0.301 & 0.282 &0.152& 0.272 & 0.043 & 63.43 &1.749\\
    & AFS~\cite{infoswap}               & 0.448 & 0.508 &0.121& 0.231 & 0.022 & 26.78 &-0.277\\
    & ReliableSwap~\cite{fslsd}         & 0.492 & 0.532 &0.126& 0.238 & 0.022 & 44.76 &-0.089\\
    
    \Xhline{.8\arrayrulewidth}
    \multirow{2}{*}{\textbf{Source-Oriented}} 
     &{FSGAN~\cite{fsgan}}   &0.338&0.403& 0.164&0.193&0.016 & 42.56 &0.354\\ 
     & {E4S~\cite{e4s}}     &0.501 &0.588& 0.118&0.262&0.028 & 52.90&0.089\\
    \Xhline{.8\arrayrulewidth}
  

    \textbf{SAMAE} & {\textbf{Ours}} & \textbf{0.578} & \textbf{0.628} & \textbf{0.108}&0.190 & \textbf{0.013} &\textbf{21.22} &\textbf{-1.326}\\
    \Xhline{2.5\arrayrulewidth}
    \end{tabular}
    \end{adjustbox}
    \vspace{-1.cm}
    
    \label{supp_table:baselines_more}
\end{table*}

\section{User Study}
For the human evaluation about criteria in the Background section in the \sloppy{manuscript (\textcolor{olive}{C1}, \textcolor{olive}{C2}, and \textcolor{olive}{C3})}, we have conducted the user studies for three aspects, magnitude of identity reflection, attribute preservation, and naturalness. We nominate the top-3 overall scored baselines from Table~\ref{supp_table:baselines_more} for the simplicity. 
We requested data from 33 subjects for these clauses. As illustrated in the table, our results exhibited the highest scores, with a significant margin over other baseline measures. 
We have also demonstrated that our method attains state-of-the-art quality in terms of human perceptual evaluation.

\begin{table}
    \caption{User study with SimSwap~\cite{simswap}, InfoSwap~\cite{infoswap}, and BlendFace~\cite{blendface} nominated as top-3 Overall scored from Table~\ref{supp_table:baselines_more}.}
    \centering
    \small
    \begin{tabular}{c|ccc}
    \Xhline{2.5\arrayrulewidth}
    Methods &  Identity $\uparrow$& Attributes $\uparrow$& Naturalness $\uparrow$\\
    \Xhline{0.8\arrayrulewidth}
    SimSwap~\cite{simswap}  &2.50&3.34&2.65\\ 
    InfoSwap~\cite{infoswap} & 2.93&3.29&2.95\\
    BlendFace~\cite{blendface} & 2.71&2.95&3.04\\
    \Xhline{0.8\arrayrulewidth}
    \textbf{Ours} & \textbf{4.18}&\textbf{3.81}&\textbf{3.97}\\

    \Xhline{2.5\arrayrulewidth}
    \end{tabular}
    
\label{supp_table:user_study}
    
\end{table}

\section{Architecture Details}
We utilize ADM~\cite{adm} U-Net architecture as the generator $G$.
The model comprises a sequence of residual layers and downsampling convolutions, followed by another sequence of residual layers featuring upsampling convolutions, with skip connections linking layers that share identical spatial dimensions.
For the $E_\textit{skin}$ architecture, we use a pretrained ResNet18~\cite{resnet} and further train the model together with the generator.
As for the discriminator, we directly adopt the architecture from the StyleGAN2~\cite{stylegan2} discriminator.

\section{Training Objectives}
The predicted image $\hat{I}$ is generated as described in the main paper:
$\hat{I} = G(I_\textit{ras}, I_\textit{p}, c_\textit{id}, c_\textit{skin})$. We devise reconstruction loss between $\hat{I}$ and $I$ with $L1$ loss and perceptual loss~\cite{vgg, perceptual},
\begin{equation}
\mathcal{L}_\textit{REC} = \mathcal{L}_\textit{L1} + \mathcal{L}_\textit{VGG}(\hat{I}, I),
\end{equation}
where  $\mathcal{L}_\textit{L1} = \mathbb{E}[|| I - \hat{I}||_1]$.
We also utilize identity loss~\cite{arcface} to maximize identity similarity: 
\begin{equation}
    \mathcal{L_\textit{ID}}  = 1-\text{cossim}(E_\textit{FR}(\hat{I}), E_\textit{FR}(I)),
\end{equation} 
where cossim refers to cosine similarity.
For improved realism of the generated images, we employ a non-saturating GAN loss~\cite{gan,stylegan2}:

\begin{equation}
\mathcal{L}_\textit{GAN}(G, E_\textit{skin}, D) = \underset{G, E_\textit{skin}}{\text{min}}\underset{D}{\text{max}}\mathbb{E}[f(D(-\hat{I}))] + \mathbb{E}[f(D(I))],
\end{equation}
where $f(u) = -\text{log}(1+\text{exp}(-u))$. 
Additionally, for the eye-gaze controllability of the model, the rendered iris keypoints image from the real image $I$ is concatenated with $I$ itself, and this pair is regarded as the real sample. 
Other cases including a real image concatenated with perturbed keypoints image or keypoints image concatenated with the generated images are regarded as fake samples.  
The total loss function is composed as a weighted sum of these loss functions:
\begin{equation}
\mathcal{L} = \lambda_\textit{REC}\mathcal{L}_\textit{REC} + \lambda_\textit{ID}\mathcal{L}_\textit{ID} + \lambda_\textit{GAN}\mathcal{L}_\textit{GAN}.
\end{equation}
For all experiments in our paper, we set hyperparameters $\lambda_\textit{REC} = \lambda_\textit{ID} = \lambda_\textit{GAN} = 1$.

\begin{table}
    \caption{Ablation study on dimensionality of $c_\textit{skin}$.}

    \centering
    \small
    \begin{adjustbox}{width=.3\linewidth}
    \begin{tabular}{l|c}
    \Xhline{2.5\arrayrulewidth}
    dim. of $c_\textit{skin}$ &  ID. Sim. $\uparrow$ \\
    \Xhline{0.8\arrayrulewidth}
    8  &0.580\\ 
    32 & 0.580\\
    64 (Ours) & 0.578\\
    128 & 0.568 \\
    512 & 0.566 \\

    \Xhline{2.5\arrayrulewidth}
    \end{tabular}
    \end{adjustbox}
    
\label{supp_table:c_dim}

\end{table}

\section{$c_\textit{skin}$ dimensional ablation}
Determining the dimensionality of $c_\textit{skin}$ is empirical, but thorough analysis establishes that a 64-dimensional representation is most suitable for our model, as depicted in Table~\ref{supp_table:c_dim}. Increasing the dimensionality of $c_\textit{skin}$ beyond 64 may lead to potential issues, such as the leakage of target identity through the portrayal of wrinkles or dimples in the target's skin, while compromising identity preservation. 

\section{Disentangled Control}

\begin{figure*}
  \centering
   \includegraphics[width=0.94\linewidth]{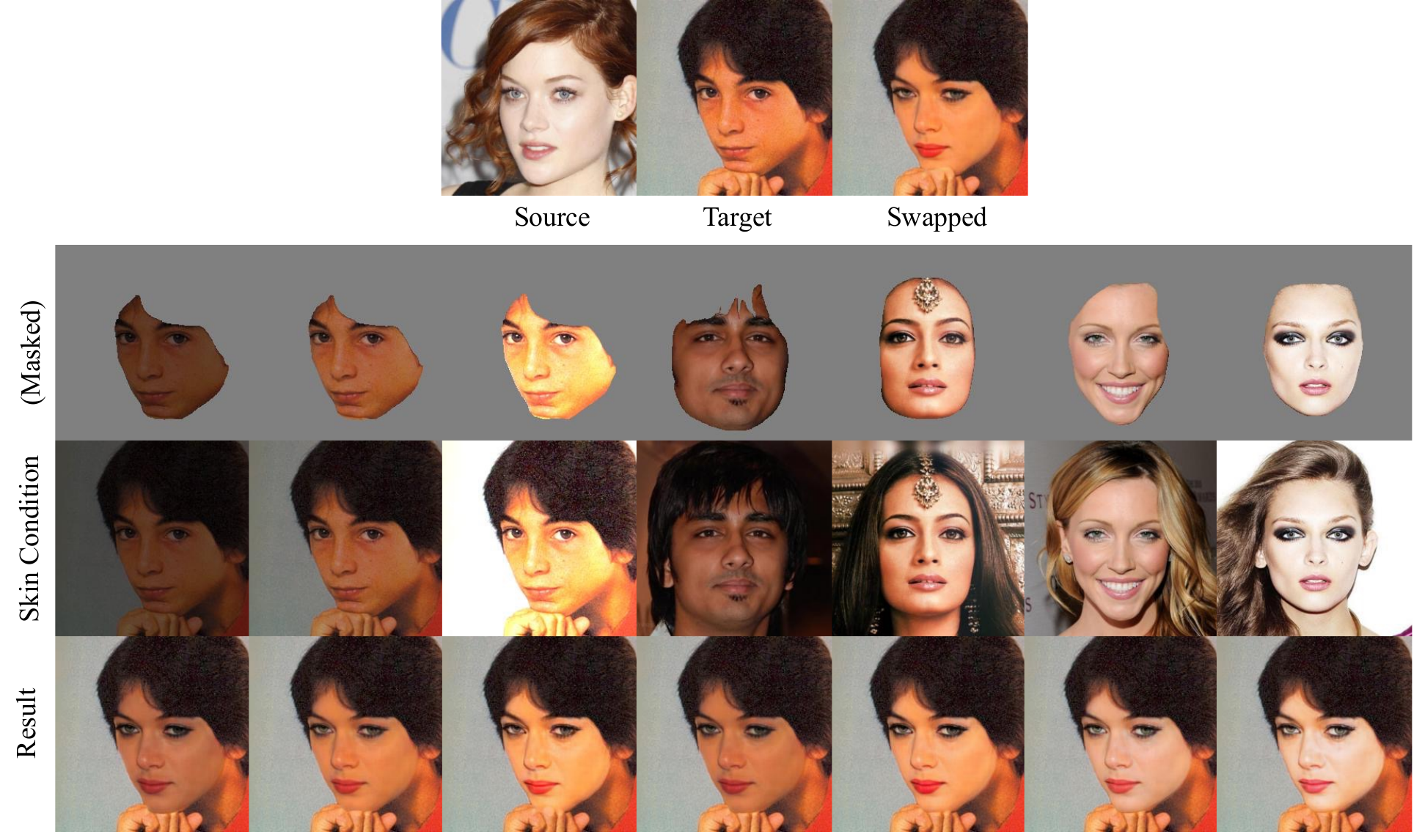}
   \caption{Skin condition image is replaced with other images or with the original target image with different brightness ((Masked), Skin Condition). The last row shows the results.}
   \label{fig:skin_cond1}
\end{figure*}

To validate the effect of $c_\textit{skin}$, we modified the skin condition image while maintaining the same source and target images. As depicted in Fig.~\ref{fig:skin_cond1}, we substituted the skin condition image with different images or altered the original target image's brightness. 
The last row of Fig.~\ref{fig:skin_cond1} displays results with varied skin color.
This experiment indicates that our approach effectively separates skin color without affecting other facial features.

\begin{figure*}
  \centering
   \includegraphics[width=0.94\linewidth]{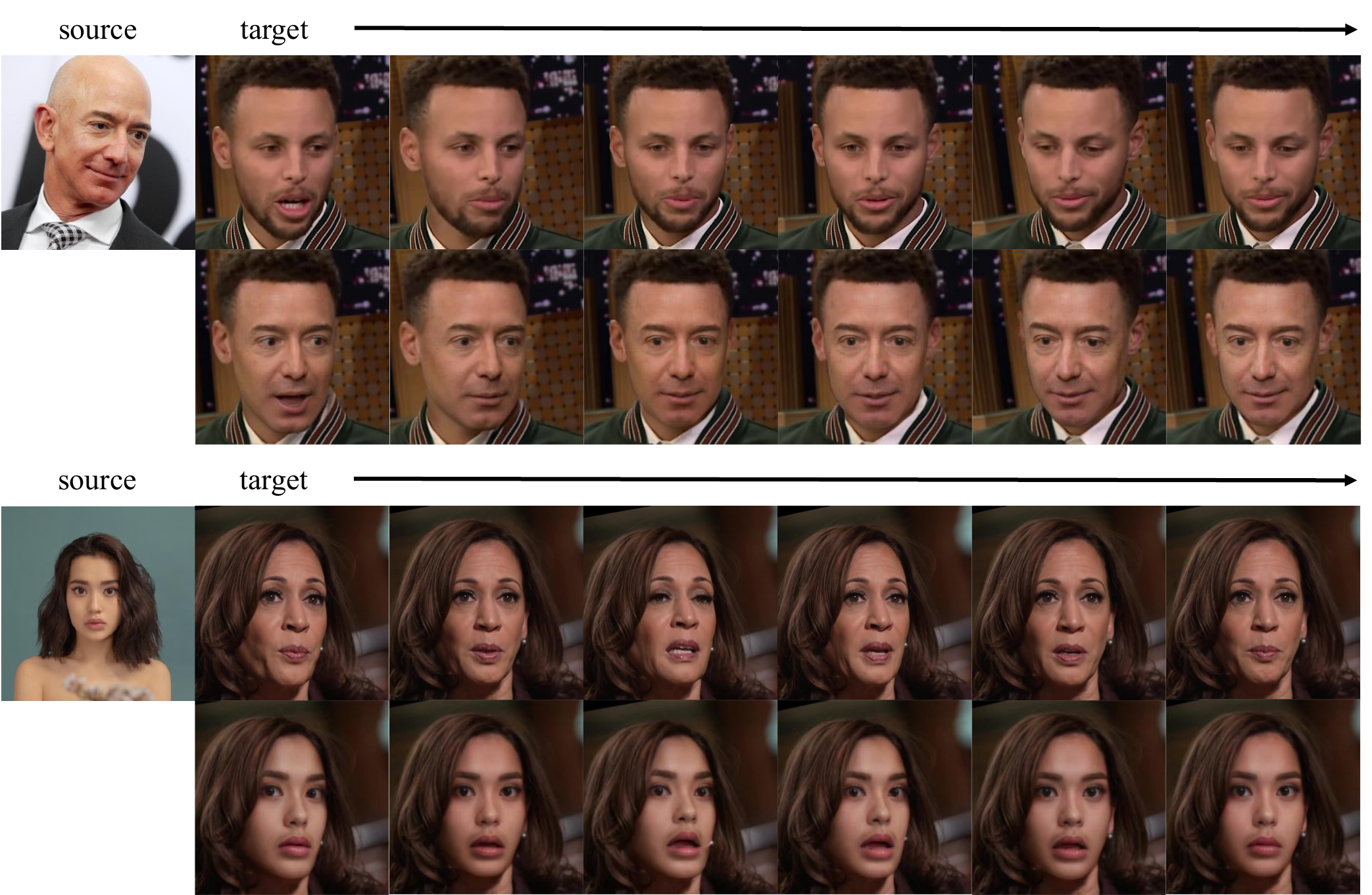}
   \caption{Controlling pose and expression with different target frames from a video.}
   \vspace{-0.7cm}
   \label{fig:pose_control}
\end{figure*}

Additionally, we evaluated the pose and expression adaptability of our model using various target frames from a video. 
The top row of Fig.~\ref{fig:pose_control} illustrates a source image alongside different target frames with varying poses and expressions.
The bottom row presents the swapped results, where the images successfully emulate the target's poses and expressions while preserving the source identity.
This demonstrates our model's capability to disentangle the poses and expressions from other facial attributes.

\section{Time-Space Complexity Comparision}
Our model has comparable number of trainable parameters and \sloppy{Multiply–Accumulate (MAC)} operations (Tab.~\ref{supp_table:complexity}).
We also tested the robustness of our method by reducing the number layers and channel sizes (\textit{Ours Small}).
Even with decreased parameter, the performance remains competitive to the original model.

\begin{table}
    \caption{Complexity comparison with ID. Sim. and Expression (Exp.) performance.}

    \centering
    \small
    \begin{adjustbox}{width=0.75\linewidth}
    \begin{tabular}{c|cccc}
    \Xhline{2.5\arrayrulewidth}
    Methods &  \# trainable parameters $\downarrow$ & MACs $\downarrow$ & ID. Sim. $\uparrow$ & Exp. $\downarrow$\\
    \Xhline{0.8\arrayrulewidth}
    SimSwap &\underline{55M}& \textbf{193B} &0.525&0.204\\ 
    InfoSwap & 203M&\underline{240B}&0.527&0.233\\
    FSLSD &163M&1.9T&0.330&0.207\\
    BlendFace & 420M&359B&0.440& \textbf{0.189} \\
    DiffSwap & 169M&22T&0.347&0.193\\
    FSGAN&200M&847B&0.338&0.193\\
    E4S & 162M&44T&0.501&0.262 \\
    \Xhline{0.8\arrayrulewidth}
    Ours-ab. & 144M & 559B & 0.477 & 0.237 \\
    \Xhline{0.8\arrayrulewidth}
    \textbf{Ours Small} &\textbf{35M}&369B&\underline{0.566}&0.193\\
    \textbf{Ours} & 144M&559B&\textbf{0.578}&\underline{0.190}\\
    

    \Xhline{2.5\arrayrulewidth}
    \end{tabular}
    \end{adjustbox}
    \vspace{-0.3cm}
    
\label{supp_table:complexity}
    
\end{table}

\section{Quantitative evaluation of Ours-ab.}
As described in the main manuscript, we demonstrate the effectiveness of self-supervised training of SAMAE (Ours) against Ours-ab. in Tab.~\ref{supp_table:complexity}.

\section{Video}
Although our model was not trained with a video dataset, it can still generate convincing swapped videos. Please refer to the attached video and image files for the results.

\section{Evaluation on FF++ benchmark}

\begin{figure*}
  \centering
   \includegraphics[width=0.65\linewidth]{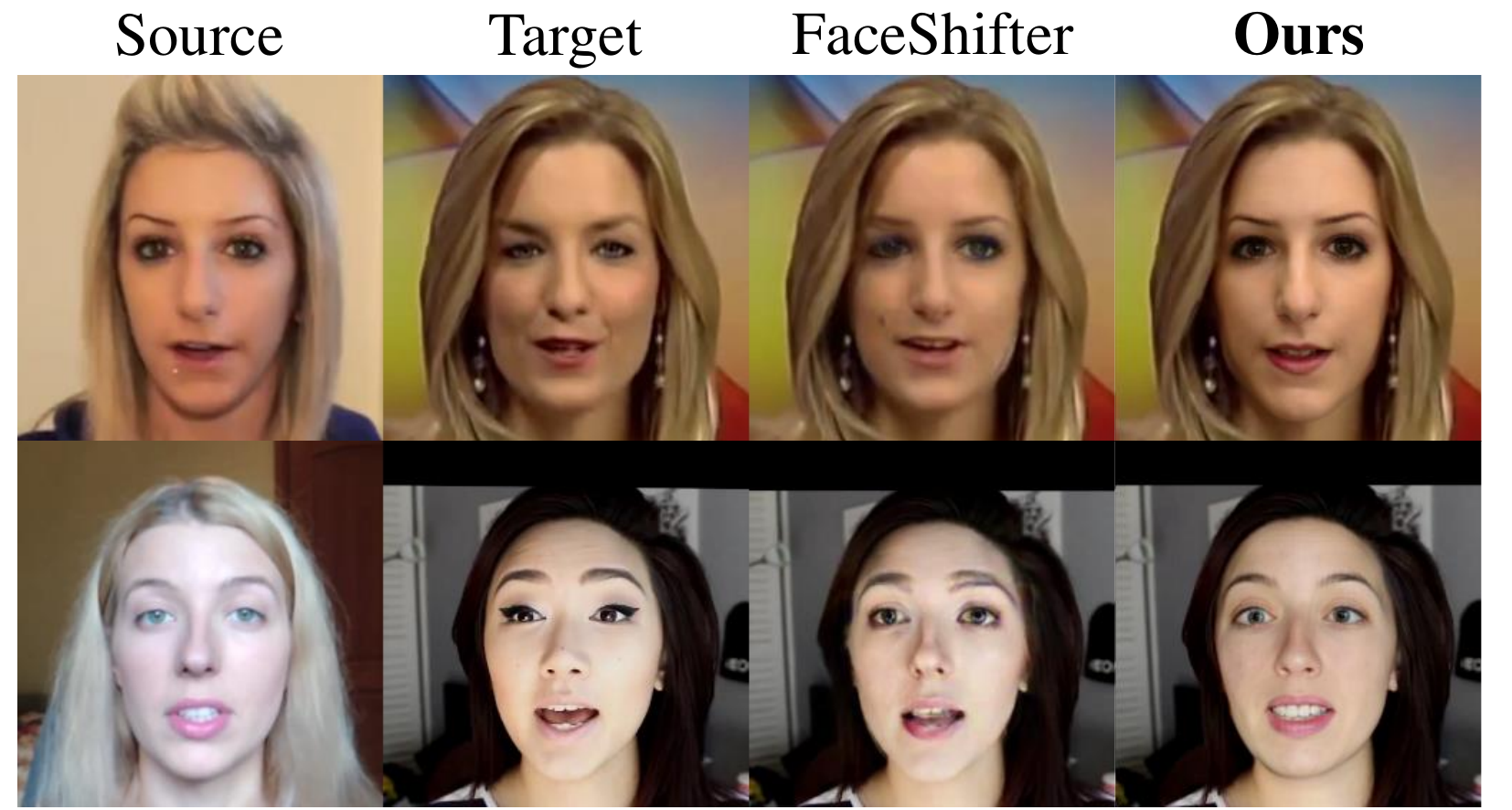}
   \caption{\text{Qualitative comparison with FaceShifter} on FF++.}
   \vspace{-0.5cm}
   \label{supp_fig:ff++}
\end{figure*}

Given the absence of official code, we perform a qualitative and quantitative comparison with FaceShifter~\cite{faceshifter} based on the provided results in FaceForensics++ (FF++)\cite{ff++}. The qualitative assessment is illustrated in Fig.\ref{supp_fig:ff++}, while the quantitative results are summarized in Table~\ref{supp_table:faceshiter}. Notably, our approach attains superior scores across all metrics with a substantial margin compared to FaceShifter on other evaluation protocols.

\begin{table}
    \caption{\text{Quantitative comparision with FaceShifter} on FF++.}

    \centering
    \small
    \begin{tabular}{l|cccc}
    \Xhline{2.5\arrayrulewidth}
    Methods &  ID. Sim. $\uparrow$ &  Shape. $\downarrow$ &  Expression $\downarrow$ &  Pose $\downarrow$ \\
    \Xhline{0.8\arrayrulewidth}
    FaceShifter &0.541&0.122&0.213&\textbf{0.019}\\
    Ours&\textbf{0.646}&\textbf{0.108}&\textbf{0.189}&\textbf{0.019}\\

    \Xhline{2.5\arrayrulewidth}

    \end{tabular}
    
\label{supp_table:faceshiter}
\end{table}

\clearpage
\section{Self-Supervised (SAMAE) vs. Target-Oriented Regime (Seesaw Game)}

In this subsection, we will discuss the pros and cons of SAMAE and the seesaw training regime. As mentioned in the manuscript, target-oriented training's reconstruction losses, such as L1 and Perceptual~\cite{vgg}, ensure comprehensive preservation of the target image's attributes, including skin color and even extreme illumination. However, this reconstruction loss acts as a double-edged sword by globally imposing these attributes on the swapped images, making it difficult to manually exclude the target's shape leakage. Conversely, our self-supervised training regime excels in preventing the target's shape leakage. However, this regime also has the drawback of being unable to guarantee preservation under harsh illumination conditions, as the $c_\textit{skin}$ is highly compressed vectorized information. This presents an ill-posed problem of decomposing the albedo and light environment from a single field of view in a single image. In future research, we believe that combining single-shot portrait relighting techniques with SelfSwapper would be a promising direction.

\section{Code Release}
We have included the implementation of our work in the attached .zip file, and we intend to release the refined code publicly in the near future.
\clearpage
{
    \small
    \bibliographystyle{splncs04}
    \bibliography{main}
}
